\documentclass{nle}

\usepackage{graphicx}
\usepackage{url}
\usepackage{color}

\title[Effective weakly supervised semantic frame induction]
      {Effective weakly supervised semantic frame induction using expression sharing in hierarchical hidden Markov models}
\author[Van de Loo and others]
       {Janneke van de Loo\textsuperscript{1}, Jort F. Gemmeke\textsuperscript{2}, Guy De Pauw\textsuperscript{1,3},
        Bart Ons\textsuperscript{2}, \and \vspace{0.3cm} Walter Daelemans\textsuperscript{1}, Hugo Van hamme\textsuperscript{2}\\       
       \textsuperscript{1}CLiPS - Computational Linguistics Group, University of Antwerp, 2000 Antwerp, Belgium\\
       	\textsuperscript{2}Department ESAT-PSI, KU Leuven, 3001 Heverlee, Belgium\\
        \textsuperscript{3}TEXTGAIN, Belgium}

\pagerange{\pageref{firstpage}--\pageref{lastpage}}
\pubyear{2016}

\begin{document}
\label{firstpage}
\maketitle








\begin{abstract}
We present a framework for the induction of semantic frames from utterances in the context of an adaptive command-and-control interface. The system is trained on an individual user's utterances and the corresponding semantic frames representing controls. During training, no prior information on the alignment between utterance segments and frame slots and values is available. In addition, semantic frames in the training data can contain information that is not expressed in the utterances. To tackle this weakly supervised classification task, we propose a framework based on Hidden Markov Models (HMMs). Structural modifications, resulting in a hierarchical HMM, and an extension called \textit{expression sharing} are introduced to minimize the amount of training time and effort required for the user.

\noindent The dataset used for the present study is {\sc patcor}, which contains commands uttered in the context of a vocally guided card game, \textit{Patience}. Experiments were carried out on orthographic and phonetic transcriptions of commands, segmented on different levels of n-gram granularity. The experimental results show positive effects of all the studied system extensions, with some effect differences between the different input representations. Moreover, evaluation experiments on held-out data with the optimal system configuration show that the extended system is able to achieve high accuracies with relatively small amounts of training data.

\end{abstract}



\section{Introduction}
\label{introduction}

The use of vocal interfaces in our daily lives is becoming
more common: we can talk to our smartphones through Siri, computers, smart-TV and other specialized domestic devices, such as Alexa and Echo. People with physical disabilities, for whom manual operation of such devices requires exhausting effort, could greatly benefit from such a hands-free control
interface. However, many people with physical disabilities additionally have
speech disorders, since motor impairments can also affect the control
of the speech articulators. This makes accurate speech recognition
very difficult. Still, case studies have also shown that, despite a
speech disorder, some users find it easier to use a speech recognizer
than a keyboard or a switch-scanning system \cite{Chang1993,Hawley2007}. 

In the ALADIN project\footnote{http://www.aladinspeech.be}, we aim to develop a speaker-dependent, adaptive
vocal interface for home automation, in which the vocabulary and command
structures are not predefined, but rather automatically induced by the system.
This allows users to address the system in an intuitive way, choosing their own commands.
The system is language independent and can adapt to regional
or pathological features of the user's speech. The vocal interface is trained in an
initial training phase in interaction with the user, and keeps
adapting to new data that are automatically collected during the
usage phase. This constant adaptation makes the system very
appropriate for people with progressive diseases. 

During the training phase, spoken commands are associated with
executed controls, which are represented as semantic frames that
encode the relevant properties of the actions. An action such as
pressing the button ``4'' on the TV remote control, associated with
the spoken command ``switch the TV to channel four'', is represented by a
frame of the type {\tt change\_channel}, containing the slots $<${\tt
  device}$>$ and $<${\tt channel}$>$ and their respective values {\tt
  TV} and {\tt 4}. 
  
A semantic frame induction engine (\textit{FramEngine}) then
looks for recurring patterns in the commands -- which may be words, morphemes 
and/or other units -- and relates
them to slots and their values in the associated semantic frames. This
induction task is weakly supervised, as there is only supervision at 
the utterance level: no relations between parts of the 
utterances and parts of the semantic frames are specified in
advance. An important requirement for the ALADIN system is that it
needs to be able to learn these relations on the basis of a small set of
training instances, since the amount of effort required from the user
to train the system should be kept to a minimum.

In previous work \cite{ons2013self}, we presented the
standard semantic frame induction system (\textit{FramEngine}) that has been developed in the
ALADIN project, and demonstrated the performance of an early
implementation of this system with non-pathological and pathological
speech input. 
The results show that the system has a
promising learning potential with small amounts of training data, but that
enhancements are needed in order to produce practically
usable accuracies for more complex utterances. Improvements can be made both in the acoustic
processing and in the way semantic frames are induced from the utterance. This paper focuses on the latter problem and studies the effect of extensions to the original Hidden Markov Model approach when trained and evaluated on the basis of {\bf transcribed} command utterances. 

Factoring out the acoustic
complexities of the task allows us to evaluate the semantic frame
induction framework in optimal conditions and observe what is minimally needed
to reliably bootstrap semantics from a signal. In this chapter, we consider
different degrees of complexity and vary the granularity of the transcription (lexical vs. sub-lexical vs supra-lexical). Experiments with transcribed data thus allow us to set an upper bound to what can be expected of semantic frame induction when it is applied to acoustic signals. In addition, using textual rather than acoustic input enables a more thorough qualitative analysis of the system's performance, since the identities of the command segments -- text segments rather than acoustic patterns -- are readily observable. In particular, the produced mappings between the command segments and the slots and values in the semantic frames can be inspected in detail. 

Our aim is to find an appropriate level of generalization: the system should not merely learn to map full utterances to full semantic frames, but rather learn associations between parts of the utterances and parts of the semantic frames and be able to make inferences about new combinations of such parts, which have not been encountered in the training data. Furthermore, we will not only focus on achieving the highest possible classification accuracy, but also on finding out which experimental conditions minimize the amount of training time needed to achieve workable results for the user. We will therefore rely extensively on learning curve experiments to evaluate the proposed techniques against the backdrop of the self-learning, adaptive command-and-control interface envisioned by the ALADIN project.

In this case study, we use a dataset of commands and semantic frames for a
voice-controlled version of the card game {\em Patience}. This is an appropriate 
application in a domestic context with an interesting level of
complexity, as the vocabulary needed to play Patience is fairly
limited, but being able to model more complex aspects such as word order, is crucial in determining the
nature of the card moves, i.e. the meaning of the commands. This makes the Patience task more complex than typical home automation tasks, such as the control of lights, heating or the television, which only require keyword spotting for successful semantic frame induction.

We will start this paper with a description of the task of semantic frame
induction in general and the standard ALADIN approach in particular 
in Section \ref{semframe}. We will describe the data for our case 
study in Section \ref{patcor}. The extensions to the architecture are presented in Section \ref{extensions}, while Section \ref{experimentalSetup} outlines the research questions 
that are addressed in this paper and present the experimental setup to
answer them. This is followed by a discussion of the
experimental results in Sections \ref{results} and
\ref{evaluation}, after which we present our
conclusions and plans for future research in section \ref{conclusion}. 

\section{Semantic Frame Induction}
\label{semframe}

The task of inducing semantic representations from utterances is well studied in the context of natural language database querying. \cite{Zettlemoyer:2005:LMS:3020336.3020416} describe an approach based on Probabilistic Combinatory Categorial Grammars to tackle the problem. Their research highlights the need to move beyond what a traditional HMM-approach is capable of. This point is also made by \cite{Chen:2011:LIN:2900423.2900560}, who describe how a semantic parser can be automatically built by observing human actions. 

The work presented in this paper differs from these research efforts in that the ALADIN approach is designed to be applicable to acoustic, as well as textual units. As such, it is more akin to research efforts in the context of spoken language 
understanding (SLU), many of which use semantic frames \cite{Wang2011}, or at least a representation that can be easily converted into such a
frame-based representation. 

Various semantic frame induction approaches have been investigated, based
on \textit{fully aligned} training data in which all the slots in the
semantic frames have been aligned with their corresponding word(s) in
the utterances. When non-hierarchical semantic representations are used in such a
\textit{supervised} context, the semantic frame induction task is
essentially a supervised sequence labeling task, akin to 
``concept tagging'', in which the words of an utterance are tagged with concepts
(slots) from the semantic representation. \cite{Hahn2011} apply a 
variety of discriminative and generative techniques to perform concept tagging of
transcribed speech corpora \cite{Bonneau2009,Mykowiecka2009,Dinarelli2009}.
Previous work in the ALADIN project similarly
applied an exemplar-based supervised concept tagging method, using
manually tagged PATCOR (cf. Section~\ref{patcor}) utterances 
as training data \cite{nlp4ita}. 

In the work presented here, we use a generative concept tagging
approach, with a lower level of supervision. In most generative
concept tagging models, hidden concept sequences are modeled with a
concept n-gram model 
and each concept state in the sequence generates a word sequence
according to another model, which \shortcite{Wang2011} call the lexicalization
model. An early generative model used for concept
tagging was the hidden semi-Markov model by \shortcite{Pieraccini1991}. 
This model was applied to the Air Travel Information
System (ATIS) dataset \cite{Hemphill1990,Dahl1994}. 
In \shortcite{Pieraccini1991}, 
the lexicalization model
was a word n-gram model, conditioned on the concept state. With n=1,
this results in a classical hidden Markov model (HMM); with n$>$1, this
is a hidden semi-Markov model (HSMM). The HMM model in the default configuration of ALADIN (Section \ref{baselineHMM}) corresponds to the n=1 version of their model. However, in our model, we use slot values as concept states, while in \shortcite{Pieraccini1991}, the concept
states correspond to slots, as in most concept tagging systems.
In most systems, the slots are first induced through a concept (=slot) tagging process, and the slot values are added
in a separate post-processing step. In ALADIN's decoding process,
on the contrary, the command units are directly tagged with slot values, which eliminates the need for additional post-processing.

The models discussed above were all applied to \textit{supervised}
concept tagging tasks. In the experiments described in this paper, such alignments 
will not be available. For concept tagging based on \textit{unaligned} data, some
generative methods based on statistical machine translation (SMT)
techniques have been used, in which the
alignment between words and concepts is explicitly modeled, 
using expectation maximization for parameter optimization \cite{Epstein1996,Pietra1997,Macherey2001}. 

It is important to note, however, that in the
experiments described in \shortcite{Epstein1996} and \shortcite{Macherey2001}, 
most words expressing concepts were replaced with class
names (such as CITY), thereby constraining their possible
alignments to concepts. Such prior class member information is also not
available in the ALADIN training situation. Since the command input in
the final ALADIN system will consist of anonymous categorical `word'
units, rather than known lexical items, no prior lexical information
can be used to constrain the alignments.

In all of the aforementioned approaches, all concepts in the
annotations were assumed to be expressed in the utterances. In the
ALADIN training situation, however, this assumption does not hold: the
semantic frames used for training are generated automatically from
actions (e.g. button presses or mouse operations), and most of the
time contain slot values that are not actually expressed in the
associated utterances. In the following subsection, we will describe
the basic ALADIN approach to perform frame decoding with a system 
trained on utterances and their associated semantic frames, very likely 
to contain redundant information.

Finally, \cite{Goldwasser:2014:LNI:2583611.2583673} describe work on learning natural language interpretations without direct supervision. While they also apply their technique to the case study of {\em solitaire}. The approach is however completely different from ours, as their goal (and the learning mechanisms to reach it) is framed in the context of learning to play the game legally, rather than to model a user's vocabulary and grammar.

\subsection{The ALADIN approach}
\label{basicAladinFramework}

\begin{figure}
\begin{center}
\includegraphics*[width=10cm]{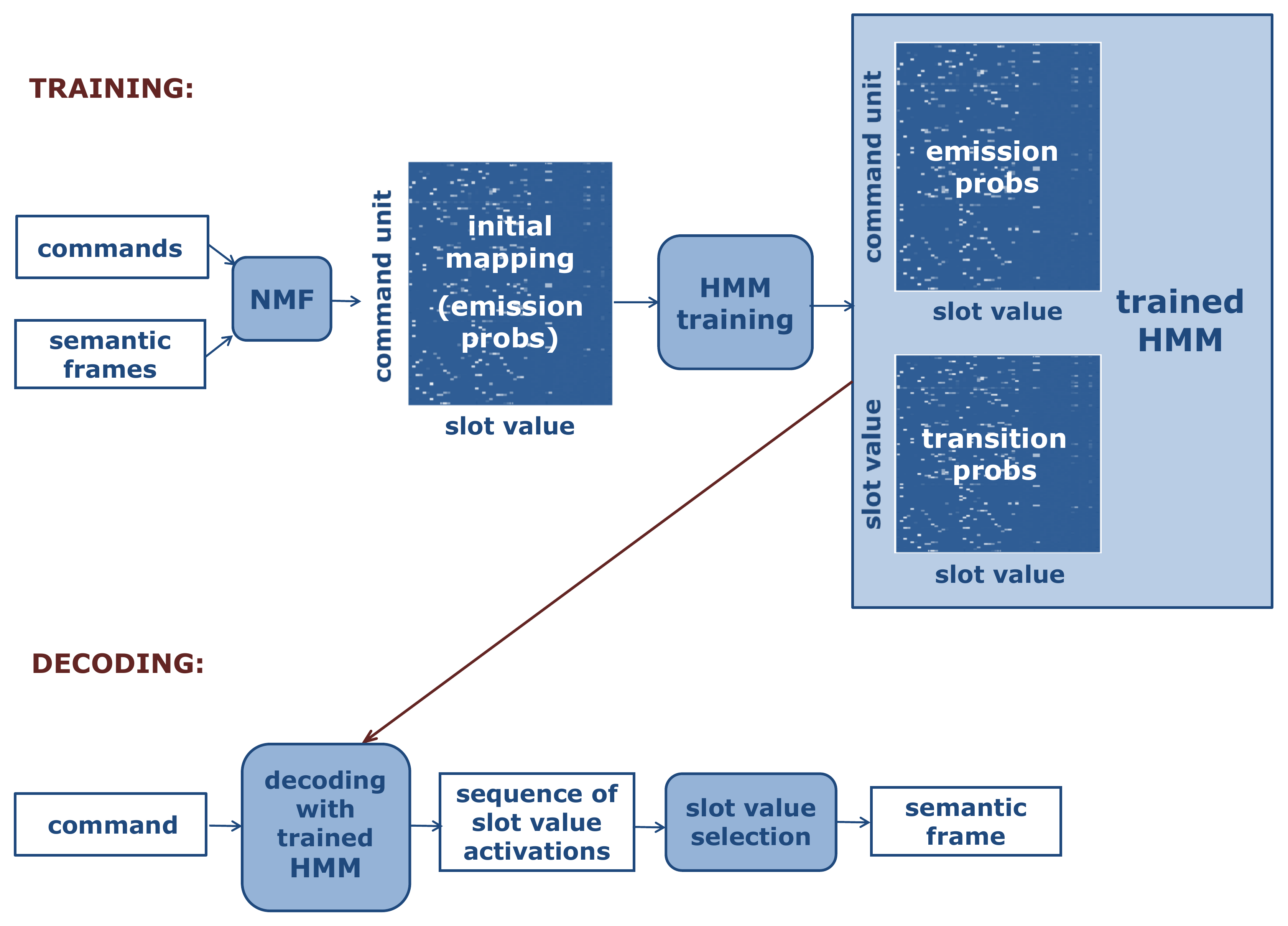}
\end{center}
\caption{The ALADIN framework.}
\label{aladinFramework}
\end{figure}

An overview of the ALADIN semantic frame induction framework is shown in
Fig.~\ref{aladinFramework}. In the {\bf training phase}, the user speaks a set
of commands, and for each 
command simultaneously executes the associated action on the device or application. The actions are automatically
converted into action frames: semantic frames in which all the
relevant properties of the action are represented in the form of slots
filled with values (see Fig. \ref{patexample}(b) for an
example). Based on this set of spoken commands and their
corresponding action frames, an HMM is trained in which the command
structures and their relations with the semantic frame structures are
modeled. HMM training is preceded by a non-negative matrix factorization (NMF)
phase, in which an initial mapping between the slot values in the
semantic frames and the observable units in the commands is produced. This
initial mapping serves as an initialization of the HMM's state
emission probability distribution.  

During {\bf decoding}, commands spoken by the user are decoded into
sequences of slot value activations, using the trained HMM. Based on
these sequences, semantic frames are generated that contain the
information on the basis of which the application can 
execute the corresponding actions. The framework will be described in
more detail in the following subsections. 

\subsubsection{Non-negative matrix factorization (NMF)}

The first step in the training process is to produce an initial
mapping between units in the commands and slot values in the semantic
frames. This is accomplished through NMF, a method that factorizes matrices
as the product of two low-rank matrices, using non-negativity constraints
\cite{lee1999}. Given a matrix V with dimensions [M x N], NMF
approximately decomposes it into a matrix W with dimensions [M x R]
and a matrix H with dimensions [R x N].

\begin{figure}
\begin{tabular}{ll}
{\bf (a)} & \includegraphics*[width=0.8\textwidth]{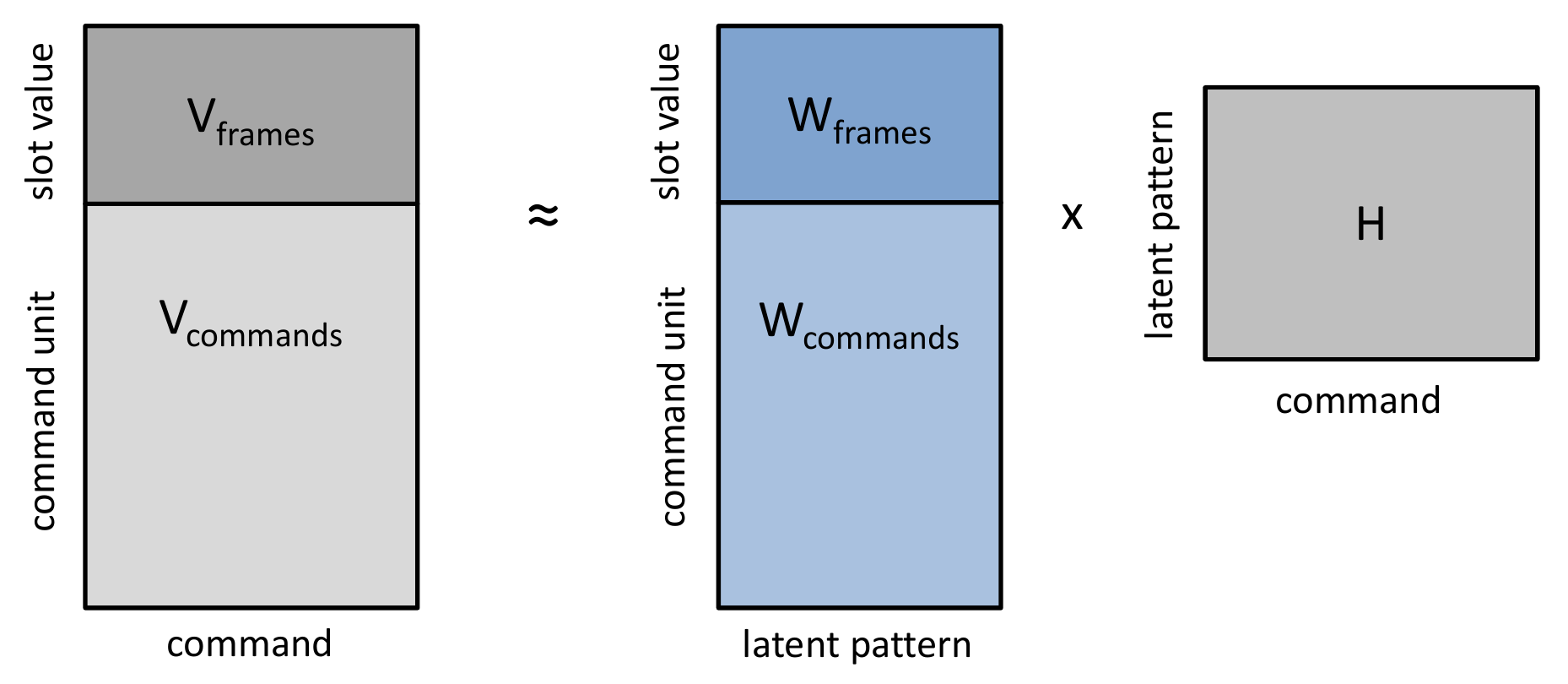}\\[8pt]
{\bf (b)} & \includegraphics*[width=0.6\textwidth]{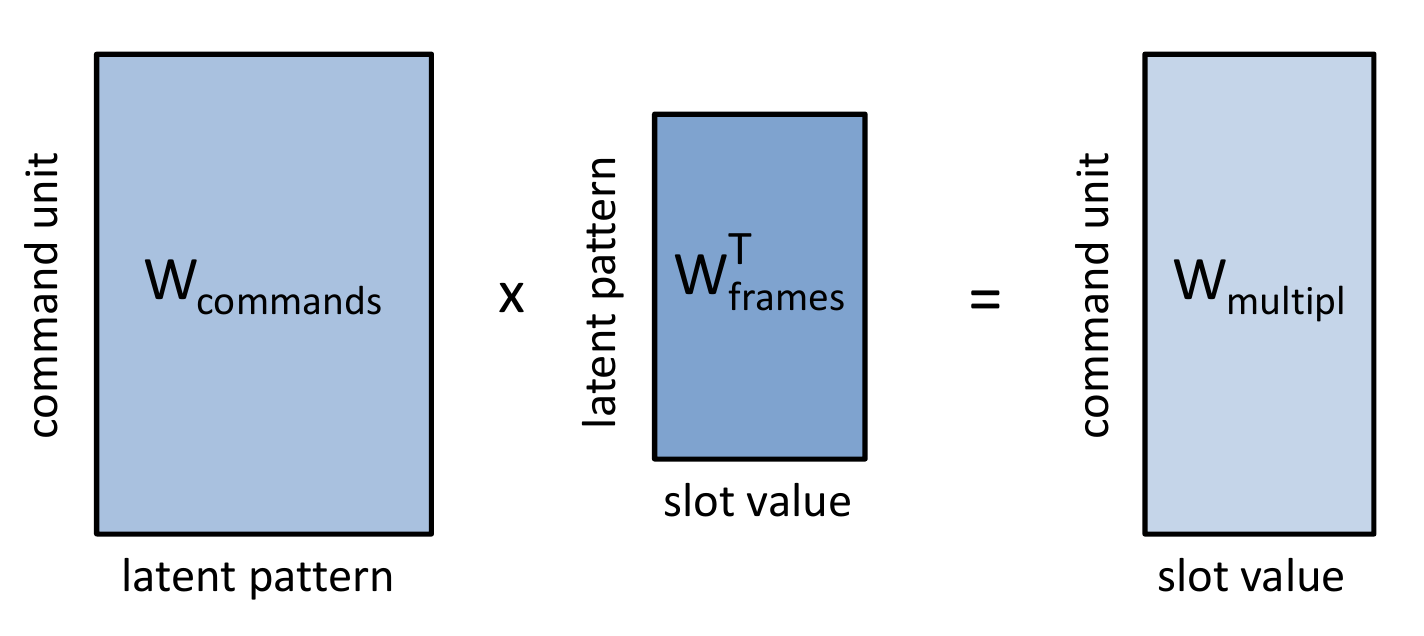}\\
\end{tabular}
\caption{The NMF process (a) and the post-multiplication of  W\textsubscript{commands} by the transpose of  W\textsubscript{frames} (b). In both (a) and (b), the input is shown on the left-hand side and the output is shown on the right-hand side of the equation.}
\label{nmfProcess}
\end{figure}

When spoken commands are used as input, NMF is used to discover
recurring acoustic patterns (e.g. word-like units) in the signal, using
the semantic frames as grounding information.  The process is depicted
in Fig. \ref{nmfProcess}(a). The input consists of two matrices: V\textsubscript{frames} and
V\textsubscript{commands}. V\textsubscript{frames} contains the frame supervision: each command column
consists of a binary vector of slot value activations, which
represents the associated semantic frame. V\textsubscript{commands} contains the
activation levels of the acoustic units that are observed in each
command: each entry contains the activation level of an acoustic unit
in a command.

The two V matrices, V\textsubscript{frames} and V\textsubscript{commands}, are vertically
concatenated, as shown in Fig. \ref{nmfProcess}(a), and decomposed into
two W matrices, W\textsubscript{frames} and W\textsubscript{commands},
and one H matrix. The columns in W\textsubscript{frames} and
W\textsubscript{commands} represent discovered latent acoustic
patterns. W\textsubscript{frames} contains the associations of these
patterns with the slot values in the semantic frames, and
W\textsubscript{commands} contains the associations with the acoustic
units observed in the commands. The matrix H contains the activation
levels of the discovered acoustic patterns in each
command. W\textsubscript{frames} constitutes the initial mapping
between the relevant command units -- which are now the discovered acoustic patterns rather than the original acoustic units -- and the slot values, as shown in
Fig. \ref{aladinFramework}. For more details regarding the NMF process
for latent acoustic pattern discovery, we refer to \shortcite{Ons2014997}
and \shortcite{hugo_hac}. 

In the work presented in this paper, textual input is used instead of
audio input, as explained in the introduction. The NMF process is the same as with audio input, as
depicted Fig. \ref{nmfProcess}(a), except that the rows in
W\textsubscript{commands} and V\textsubscript{commands} represent
textual units instead of acoustic units, i.e. word or phoneme n-grams. 
The recurring patterns that are discovered by NMF are not used in our
experiments with textual input, because the textual units themselves
are the relevant units that should be associated with slot values in
the semantic frames. Therefore, we post-multiply W\textsubscript{commands} by the transpose of 
W\textsubscript{frames}, as depicted in Fig. \ref{nmfProcess}(b). 
This results in a matrix W\textsubscript{multipl}, with rows representing textual command units and 
columns representing slot values in the semantic frames.
In our experiments, W\textsubscript{multipl} is used as the initial mapping between slot values and command units, which is depicted in Fig. \ref{aladinFramework} as the result of the NMF process.

\subsubsection{Baseline HMM}
\label{baselineHMM}

In HMMs, observed sequences are assumed to be generated by an
underlying sequence of hidden states. In the HMMs that are used in the
ALADIN framework, the commands are the observed sequences (be it
acoustic or textual). In the experiments presented in this paper, with
textual command input, they are sequences of word n-grams or phoneme
n-grams. 

\begin{figure}
\begin{center}
\includegraphics*[width=13cm]{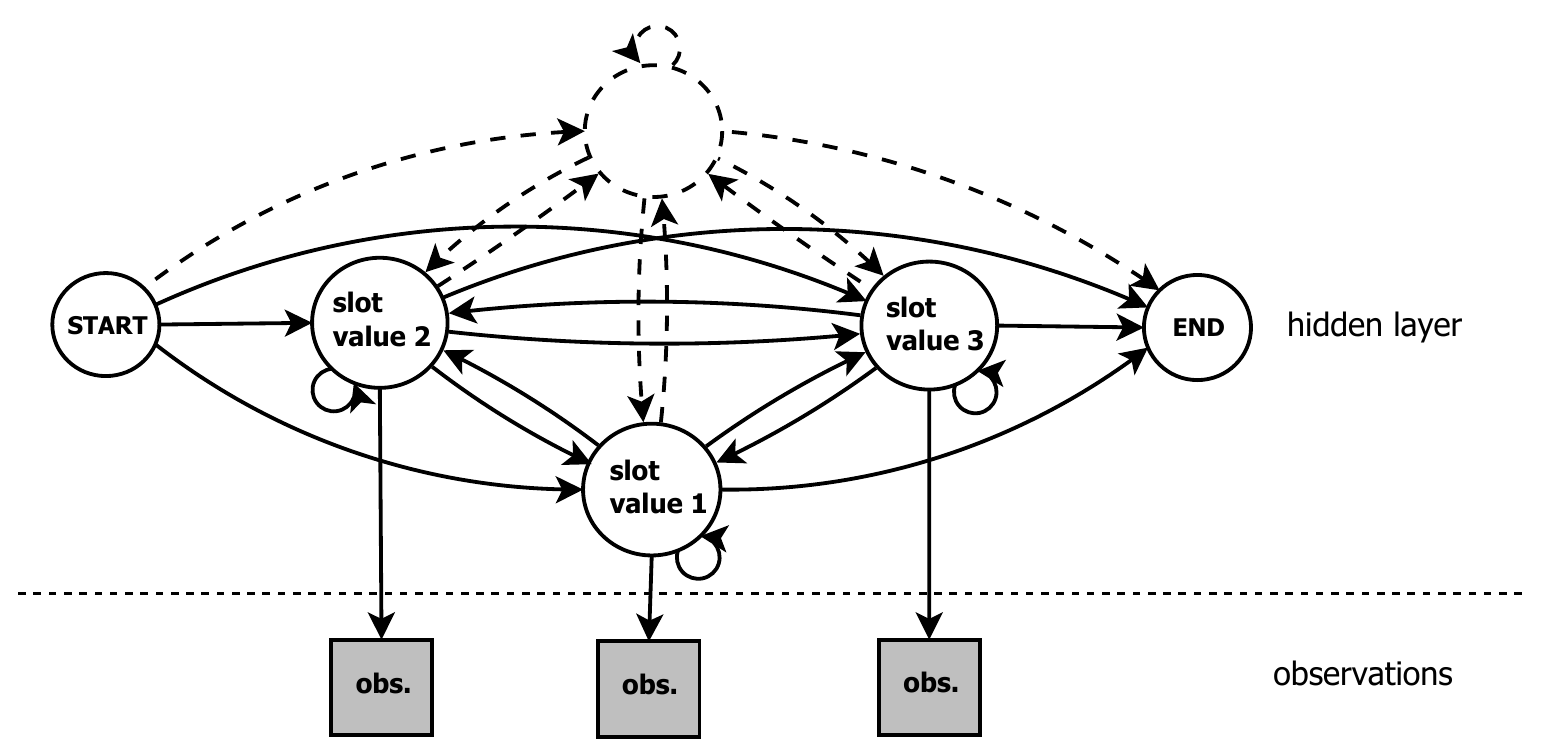}
\end{center}
\caption{The basic HMM structure.}
\label{basicHMM}
\end{figure}

The hidden states in the HMM are the slot values in the semantic frames. 
The basic HMM structure is depicted in Fig. \ref{basicHMM}.
This figure shows the HMM of one semantic frame type. In an
application where multiple semantic frame types are used, several of
these HMMs are connected in parallel; one for each frame type. The
slot value states in the single-frame HMM are almost fully
connected. The only transitions that are prohibited, are transitions
between slot values that belong to the same slot (apart from
self-transitions to the same slot value, which are allowed), and
transitions involving slot values that do not occur in the training
data. The initial non-zero transition probabilities are uniformly
distributed. The transition probability distribution can be
represented as a state-by-state matrix of probabilities -- in this
case, a slot-value-by-slot-value matrix (see Fig.
\ref{aladinFramework}).

The transition and emission distributions of the states are the HMM
parameters, which are trained in an iterative procedure using the
Baum-Welch algorithm \cite{Baum1972}. This algorithm, which is a specific
version of the expectation-maximization (EM) algorithm \cite{Dempster1977}, 
iteratively alternates between an expectation step
(E-step) and a maximization step (M-step). In the E-step, expected
state occupancy and transition counts are computed based on the
current HMM parameters and the observed sequences; in the M-step, the
HMM parameters are updated based on the counts. In ALADIN's HMM
training procedure, the semantic frame supervision is used at the end
of each E-step: the expected occupancy counts of states (slot
values) that do not occur in a given utterance according to the
corresponding semantic frame are set to zero, followed by
re-normalization. 

\subsubsection{Decoding}
\label{decoding}

In the decoding phase, the trained HMM is used to decode a command
into a sequence of slot values, which is subsequently converted into a
semantic frame. First, unknown command units, which have not occurred in the training data, 
are mapped to the most similar known unit from the
training data using ADAPT \cite{Elffers2005}. ADAPT is a dynamic programming
algorithm that computes the minimum edit distance between two strings
of phonetic symbols, based on articulatory features.

The Viterbi algorithm \cite{Viterbi1967} is then used to find the optimal path through
the slot value states in the trained HMM, given the command. 
Since this algorithm produces a single optimal path, which
can only include slot values of a single frame type, the frame type is
implicitly selected. However, it is possible that multiple slot values
for a single slot occur in the resulting slot value sequence (only
{\em direct} transitions between slot values within the same slot are
not allowed in the HMM). In order to select the most probable slot
value for each slot, the posterior probabilities of the slot values in
the sequence, given the emission probability distribution, are
used. These posterior probabilities are accumulated across the
sequence, and for each slot, the slot value with the highest total
probability is selected. A slot is filled with the selected slot value
if its total posterior probability exceeds a certain threshold.

\section{Patience dataset PATCOR}
\label{patcor}

\begin{figure}
\begin{center}
\begin{tabular}{c}\hline
Command: {\em Leg de klaveren boer op de rode koningin}\\
(English: {\em Put the jack of clubs on the red queen})\\\\
\includegraphics*[width=6cm]{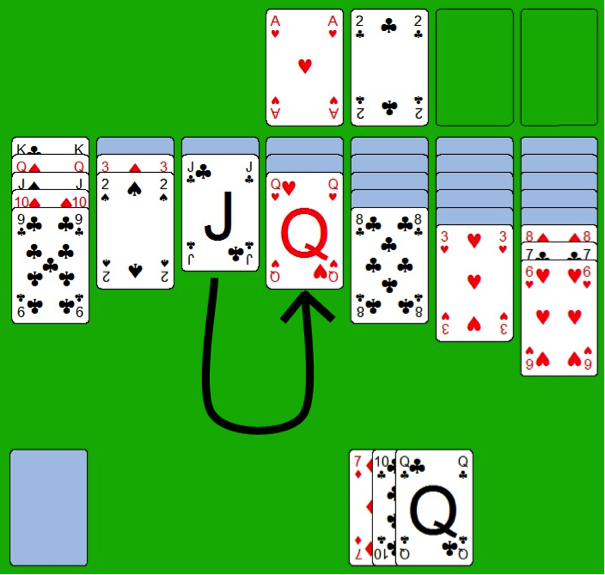}\\
{\bf (a) Command and corresponding action in the playing field}
\end{tabular}

\begin{tabular}{lcc}\hline\hline
\multicolumn{3}{c}{Frame type: {\tt movecard}}\\
     &  Automatic & Oracle\\
Slot &	Slot value & Slot value\\\hline
$<$from\_suit$>$&	c &	c \\
$<$from\_value$>$	& 11 & 11 \\
$<$from\_foundation$>$&	-&	-\\
$<$from\_column$>$&	3 &	-\\
$<$from\_hand$>$&	-&-\\
$<$target\_suit$>$&	h & h,d \\
$<$target\_value$>$&	12 &	12 \\
$<$target\_foundation$>$&	-&	-\\
$<$target\_column$>$&	4 &	-\\\hline
\multicolumn{3}{c}{{\bf (b) automatic and oracle frames}}
\end{tabular}
\end{center}
\caption{Example of a Patience command in PATCOR, the corresponding action on the playing field (a) and the content of of the automatically generated frame and the oracle frame that was added manually (b).}
\label{patexample}
\end{figure}

The experiments presented in this paper use a vocally guided Patience
game as a case study. Patience (also known as Solitaire) is one of the most well-known
single-player card games. The playing field (cf. Fig.
\ref{patexample}) consists of seven columns, four foundation
stacks (top) and the remainder of the deck, called the hand
(bottom). The aim of the game is to move all the cards from the hand
and the seven columns to the foundation stacks, through a series of
manipulations, in which consecutive cards of alternating colors can be
stacked on the columns and consecutive cards of the same suit are
placed on the foundation stacks.

For our experiments, we will make use of PATCOR, a dataset containing
recordings of nine speakers playing Patience. In total,
PATCOR contains over 3,000 spoken commands, 
supplemented with command transcriptions, corresponding semantic
frames, and representations of game states between the moves. The
language of the spoken commands is Belgian Dutch, and the speech is
non-pathological. 

The speakers' ages range between 22 and 73, and the first eight
speakers were balanced for gender and education level. With these eight
speakers, around 250 utterances were recorded per speaker. In addition, a larger dataset of over 1,000
utterances was recorded with a ninth speaker, which we will use as our
final means of evaluation on held-out data (cf. Section
\ref{evaluation}). More details about the data set and the command structures that were used
by the speakers are described in \shortcite{nlp4ita}.

\begin{table}
\caption{PATCOR example transcriptions for ``{\em zwarte drie op rooie vier}''
({\em black three on red four}). ``\_'' indicates a word boundary.}
\begin{center}
\begin{tabular}{cc}\hline\hline
Orthographic &	Phonemic\\\hline
zwarte drie op rooie vier  &	zwArt@\_dri\_Op\_roj@\_vir\\\hline
\end{tabular}
\end{center}

\label{transcripts}
\end{table}

\subsection*{Transcriptions and action frames}
\label{actionFrames}

The recorded commands were orthographically transcribed by the first author. 
The orthograpic transcriptions were then converted to phonemic transcriptions, 
using a pronunciation lexicon with only one pronunciation variant per word. The pronunciation lexicon was based on
the lexicon of the Spoken Dutch Corpus (CGN, \shortcite{Oostdijk2000}), 
from which the single pronunciation variants were
selected manually. Words not present in the pronunciation lexicon were
added manually. The phoneme alphabet used for the transcriptions is
YAPA (cf. \shortcite{Mertens1998}), as exemplified in Table \ref{transcripts}.  

The commands were also annotated with their semantic representations in the
form of {\em action frames}. The action frames in PATCOR are representations
of Patience moves, specifying the type of move - the frame type - and
a set of attributes in the form of {\em slots} that can be filled with
{\em values}. The slots and their values specify certain properties of
the move, such as the position of the card that is moved and the
position that it is moved to.

PATCOR has two frame types: {\tt dealcard} and {\tt movecard}. The two frame types 
and their associated slots and slot values (if any) are shown in Fig.
\ref{patframes}. The {\tt dealcard} frame has no slots; it simply represents
the action of dealing a new hand and needs no extra attributes. The
{\tt movecard} frame represents a card move from one position to another,
and has nine slots. The first five slots pertain to the card that is
moved (the {\tt from} slots) and the other four pertain to the card or
position that it is moved to (the  {\tt target} slots). Cards are specified
in terms of suits ({\tt h} for hearts,  {\tt s} for spades,  {\tt d} for diamonds and  {\tt c}
for clubs) and values ({\tt 1} to  {\tt 13}, representing ace to
king). Card positions are also specified in terms of three areas on the playing field:
the columns ({\tt 1} to  {\tt 7}) in the middle, the foundation stacks ({\tt 1} to  {\tt 4}) at
the top, and the hand at the bottom (cf. Fig. \ref{patexample}(a)). 

Each command in PATCOR has two action frames associated with it: an
{\em automatic frame} and an {\em oracle frame}. An example of a command, its
associated move and its two action frames is shown in
Fig. \ref{patexample}. The automatic frame and the oracle frame both
have the same frame type and slots, 
but the slot values that are filled in differ. The automatic frame was 
generated during the Patience game through
the move that was performed by the experimenter. In this frame, all
slot values that apply to the performed move, are filled in (see Fig.
\ref{patexample}(b)). These are all the relevant properties of the
move, which speakers {\em might} refer to in their commands. The automatic
frame is therefore usually overspecified, i.e. containing redundant information
not expressed in the command. 

The oracle frame, on the other hand, was added manually, and
represents the actual content of the command that was spoken (see
Fig. \ref{patexample}(c)). This means that only the slots that
the command actually refers to, are filled in. In Fig.~\ref{patexample}(c), 
for instance, the card positions are not filled
in, because they are not mentioned in the command. In addition, the
oracle frame may include multiple slot values for a single slot, in
cases where the command is ambiguous. In the example in
Fig. \ref{patexample}, the word `red' is ambiguous as to the value of
the slot $<${\tt target\_suit}$>$: it can be either hearts (h) or diamonds (d). In
such cases, the oracle frame includes all slot values that are
possible according to the command; in this case, both  {\tt h} and  {\tt d} are
included in the slot $<${\tt target\_suit}$>$.

In the experiments described below the automatic frames will be used to train the systems. The manually created oracle frames will function as gold-standard reference points against which we can evaluate.

\begin{figure}
\begin{center}
\begin{tabular}{lllp{1cm}ll}\cline{1-3}
\multicolumn{3}{c}{\bf Frame type: {\tt movecard}} \\
\multicolumn{2}{l}{\bf Slot}	& {\bf Slot values} \\\cline{1-3}
$<$from\_suit$>$&(FS)&	h,d,s,c \\
$<$from\_value$>$&(FV)&	1-13\\\cline{5-6}
$<$from\_foundation$>$&(FF)&	1-4 & & \multicolumn{2}{c}{\bf Frame type: {\tt dealcard}}\\
$<$from\_column$>$&(FC)&	1-7 & & {\bf Slot}	& {\bf Slot values}\\\cline{5-6}
$<$from\_hand$>$&(FH)&	1 & & - &-\\\cline{5-6}
$<$target\_suit$>$&(TS)&	h,d,s,c\\
$<$target\_value$>$&(TV)&	1-13\\
$<$target\_foundation$>$&(TF)&	1-4\\
$<$target\_column$>$&(TC)&	1-7\\\cline{1-3}
\end{tabular}
\end{center}
\caption{The two frame types in PATCOR, including their slots and the
  possible slot values. The full slot names are in angle brackets; the abbreviated slot names are in round brackets. The frame type {\tt dealcard} does not have any
  slots.}
\label{patframes}
\end{figure}

\section{System extensions}
\label{extensions}

The architecture described in Section \ref{basicAladinFramework} provides a full semantic
frame induction framework, enabling training and decoding with both
textual and acoustic command input. However, there is much room for improvement of this
basic system. In this section, we present some system extensions under consideration in this paper: 
two enrichments to the HMM structure and a novel technique called {\em expression sharing}. We will discuss these in the following subsections.

\begin{figure}
\begin{center}
\includegraphics*[width=12cm]{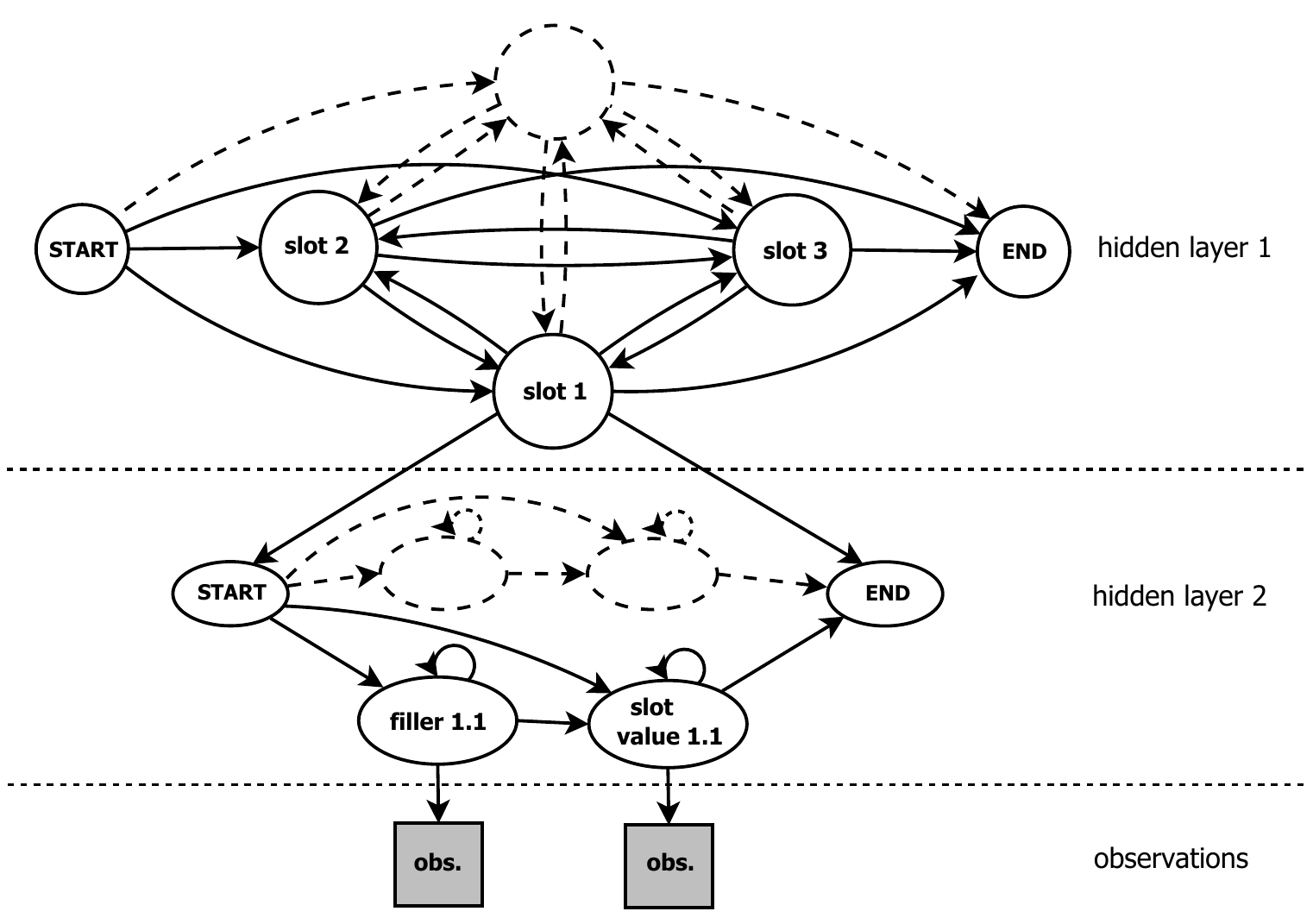}
\end{center}
\caption{The modified, hierarchical HMM structure, which includes an extra hidden layer and filler states. }
\label{hierarchicalHMM}
\end{figure}

\subsection{Slot-based transition probability sharing}
\label{transitionSharing}

In the basic HMM, the hidden layer only represents slot values; these
are the values that have to be induced by the system in order to fill
in a complete semantic frame. In most commands, however, the
underlying command structure is better defined in terms of slot
sequences than in terms of sequences of slot values; transition
probabilities hold between slots rather than individual slot
values. For instance, the transition probability between the slots
$<${\tt target\_suit}$>$ and $<${\tt target\_value}$>$ should be independent of their
specific slot values. This intuition has been implemented in ALADIN's
HMM structure by {\em sharing}, or equalizing, the transition
probabilities between all pairs of slot values belonging to a
particular pair of slots.

This introduces an extra layer in the HMM, resulting in a \textit{hierarchical} HMM (HHMM), as depicted in
Fig.~\ref{hierarchicalHMM}: the highest-level hidden layer is now a
layer of slot states, where each slot state models a sequence of acoustic events corresponding to at least a word. Each slot state has multiple sub-HMMs that
model its different slot values. The states in the slot value sub-HMMs
generate observations (command units); slot value states generate
command units expressing a specific slot value, and filler states 
generate so-called {\em filler units} (cf. Section \ref{fillerStates}).

The gain from this hierarchical architecture is a reduction in the number of transition probabilities to be estimated. Without hierarchy, each slot value can have an arbirary transition probability to the next slot value. In an HHMM, all transitions pass through the non-emitting {\em START} and {\em END} states in layer~2 of Fig.~\ref{hierarchicalHMM}, hence factorising the full transition matrix into the outer product of two vectors. Sharing HMM parameters in this way reduces the number of parameters to be learned, which should reduce the amount of training data needed. During training, transition probability sharing is carried out after the M-step in each Baum-Welch training iteration, by averaging the re-estimated
transition probabilities across shared transitions before normalizing
them. Slot values that do not occur in the training set are excluded
from sharing.  

The HHMM also provides the framework for sharing emission densities, again with the goal of reducung the number of model parameters and hence the training data requirements. Two forms of emission tying are exploited: sharing of {\em filler states} and {\em expression sharing}.

\subsection{Filler states}
\label{fillerStates}

These filler states are introduced in order to deal with
command units that do not express specific slot
values, for instance function words such as determiners or
prepositions, and interjections such as `uh' (`erm') and `nee' (`o'). Many of these
filler units can serve as signal words
that indicate certain slot expressions before or after them, for
instance in the case of prepositions. In our
framework, the filler states are associated with specific slot value
states: each slot value state is preceded by a dedicated filler state,
which can optionally be skipped.  

The filler states have a shared initial emission probability
distribution. This initial distribution is produced by adding an extra
`filler unit' column to the matrix W\textsubscript{frames} in NMF,
which is activated for all commands. In the HMM training phase, the
emission probability distributions of the filler states can optionally
be shared. 

\subsection{Expression sharing}
\label{expressionSharing}

In many applications, there are sets of slot values that are very
likely to be expressed by the same words. For instance in the Patience
application, we can assume that the slot values in the slot
$<${\tt from\_suit}$>$ are expressed by the same words as the slot values in
the slot $<${\tt target\_suit}$>$,  e.g. by the words `hearts', `spades',
`clubs' and `diamonds' in English. In traditional approaches, this property
is typically not considered during semantic frame slot filling. 
In this paper, we introduce {\em expression
sharing}, a novel technique to incorporate this knowledge
in the system. This is done by {\em sharing} the associations of these slot values with
observed units in the commands. The
sets of slots that share their slot value expressions 
are called {\em shared expression
sets}. Expression sharing can also reduce the amount of
training data needed, since it decreases the number of associations
between slot values and command units that have to be learned by the
system.  

In many cases, the shared expression sets that are defined, are sets of
slots that are essentially specific instances of a more general slot
type. For instance, the slots $<${\tt from\_suit}$>$ and $<${\tt target\_suit}$>$
can be regarded as instances of a more general slot type
$<${\tt suit}$>$. Expression sharing is therefore akin to the concept of
discerning different slot types in the semantic frame definitions,
such as the slot types `City' and `Date' in the Air Travel Information
System (ATIS) domain \cite{Hemphill1990,Dahl1994}.

In the ALADIN framework, expression sharing can be applied at two
different stages in the training process: during the NMF phase and during the
HMM training phase. In the NMF phase, expression sharing is applied as
follows: when a slot value that is part of a shared expression set is
encountered in a training instance, the other corresponding slot
values in the shared expression set are activated as well in that
training instance. For example, if an instance's semantic frame
contains the slot value $<${\tt from\_suit=h}$>$, the slot value
$<${\tt target\_suit=h}$>$ is also activated in that instance. A complete example is shown in Table \ref{nmfSharingExample}.

\begin{table}
\begin{tabular}{cc}\hline\hline
  \multicolumn{2}{c}{{\bf Command:} {\em harten acht op schoppen negen}}\\ 
\multicolumn{2}{c}{({\bf English: }{\em eight of hearts on nine of spades})}\\
\hline
\textbf{Original frame}  & \textbf{Additional frame}\\
\textbf{supervision} & \textbf{supervision}\\ \hline
FS=h & TS=h \\
FV=8 & TV=8	\\
FC=2 & TC=2	\\
TS=s & FS=s	\\
TV=9 & FV=9	\\
TC=4 & FC=4	\\
\hline
\end{tabular}
\caption{Example of expression sharing in the NMF phase: the slot values in the right column are added to the instance's frame supervision.}
\label{nmfSharingExample}
\end{table}

In the HMM training phase, expression sharing is applied by sharing
the emission probability distributions of corresponding slot values in
a shared expression set. After the M-step in each Baum-Welch pass, the
re-estimated emission probability distributions are averaged across the
corresponding slot value states (for instance, across the states
$<${\tt from\_suit=h}$>$ and $<${\tt target\_suit=h}$>$), before they are normalized.
 
Expression sharing in the HMM training phase is not only applied to
sets of corresponding slot value states, but also to sets of filler
states. Two options were implemented regarding filler state expression
sharing. The first option is to share the emission probabilities
across {\em all} filler states, resulting in one single filler state
emission probability distribution. The second option is to share the
filler state emission probability distributions slot-wise, which means
that the emissions are shared among filler states that belong to the
same slot. For instance, the emissions of the filler states
associated with the slot values $<${\tt from\_suit=h}$>$,
$<${\tt from\_suit=s}$>$, $<${\tt from\_suit=c}$>$ and $<${\tt from\_suit=d}$>$ are
shared, resulting in one single $<${\tt from\_suit}$>$ filler state emission
probability distribution. The use of slot specific filler state
emissions is similar to the use of slot specific preamble and
postamble states in \shortcite{Wang2006};
they can serve as contextual clues for identifying the slot. 

We can expect expression sharing to be a powerful extension to traditional HMM-driven semantic frame induction. It can typically be applied when concepts that need to be induced, are subtypes of a more general concept (or are used in different contexts). In the context of ATIS  \cite{Hemphill1990}, for example, departure city and destination city are both subtypes of a more general concept  'city'. Expression sharing would enable the discovery of this property. While expression sharing is able to solve a number of issues, it does not enable processing quantifiers or the induction of deep hierarchic concept spaces.

\section{Experimental Setup}
\label{experimentalSetup}

In this paper, we investigate the effect of the system extensions
discussed in the previous section on the system's semantic frame induction capabilities. We focus on the
following research questions:

\begin{enumerate}
\item{Do transition probability sharing and expression sharing have
    the expected positive effect on learning speed, and how large are
    the effects of the different sharing types?}
\item{Do these extensions introduce specific decoding errors?}
\item{How does the introduction of filler states affect the semantic 
frame induction performance and what effect do the different types of
    filler state emission probability sharing have?}
\end{enumerate}

\subsection{System parameters}
\label{syspars}
In order to investigate these effects in controlled conditions, we
perform exhaustive experiments with four system variables, based on 
the extensions discussed in Section~\ref{extensions}:

\vspace{0.2cm}
\noindent\begin{tabular}{ll}\hline\hline
{\bf Parameter} & {\bf Values}\\\hline 
Filler states &	none, non-shared, all-shared, slot-shared\\
T (transition) sharing &	true, false\\
E (expression) sharing in NMF &	true, false\\
E (expression) sharing in HMM &	true, false\\\hline 
\end{tabular}
\vspace{0.2cm}

\noindent The parameter `filler states' has four
possible values: there can be no fillers (`none'), fillers without
emission sharing (`non-shared'), fillers that all share their emission
probability distributions (`all-shared'), or fillers that share their
emission probability distributions slot-wise (`slot-shared'), which
means that the distributions of fillers that belong to the same slot
are shared. The other three parameters are booleans. T-sharing (transition-sharing) means
that the transition probabilities are shared slot-wise, as discussed
in Section \ref{transitionSharing} and depicted in
Fig. \ref{hierarchicalHMM}.  The two remaining parameters are two
different types of expression sharing applied to slot value states in respectively
the NMF phase and the HMM phase.

\subsection{Decoding methods}

Apart from the parameter variation listed in Section \ref{syspars}, we also experimented with two decoding methods: NMF decoding and HMM decoding. HMM decoding is the decoding method that was described in Section \ref{decoding}, using the trained HMM. NMF decoding, on the other hand, is a baseline decoding method in which only NMF is used. This decoding method does not model any information about the temporal ordering of the command units. The matrix with the associations between slot values and command units, which has been produced in the NMF training phase, is used to convert the sequence of command units into a sequence of slot value activations (slot value probability distributions). For each slot value, all activations across the whole sequence are accumulated, and for each slot in each frame, the slot value with the highest accumulated activation is selected, if that activation exceeds a certain threshold. Since this method can result in the selection of slot values from different frames, the frame with the highest accumulated probability mass is selected.

\subsection{Command input}

We also vary the input type during our experiments, to study the 
effect of command unit granularity on the performance of the frame induction approach. 
In the experiments reported here, we only use phonemic gold-standard
transcriptions (Table \ref{transcripts} on the right).
Different segmentations of the transcriptions are used. The
transcriptions were segmented into word unigrams, word bigrams,
phoneme unigrams or phoneme bigrams, as exemplified in the following
example: 

\vspace{0.2cm}
\noindent\begin{tabular}{lp{9cm}}
Orthographic: & {\em zwarte drie op rode vier} (black three on red four)\\
Word unigrams: & /zwArt@/ /dri/ /Op/ /roj@/ /vir/\\
Word bigrams: & {\small /+\_zwArt@/ /zwArt@\_dri/ /dri\_Op/ /Op\_roj@//roj@\_vir/ /vir\_+/}\\
Phoneme unigrams: & {\small/z/ /w/ /A/ /r/ /t/ /@/ /d/ /r/ /i/ /O/ /p/ /r/ /o/ /j/ /@/ /v/ /i/ /r/}\\
Phoneme bigrams: & {\small/+z/ /zw/ /wA/ /Ar/ /rt/ /t@/ /@d/ /dr/ /ri/ /iO/ /Op/ /pr/ /ro/ /oj/ /j@/ /@v/ /vi/ /ir/ /r+/}\\
\end{tabular}
\vspace{0.2cm}

\noindent Note that with phoneme-based input, the word boundaries are
omitted, while with word-based input, they are preserved. For the
formation of bigrams, the `+' sign was used as an extra command unit
at the beginning and the end of the utterance. 

\subsection{General setup}

Each configuration, with its unique combination of parameter
values and input type, was tested with the data of the first eight speakers in PATCOR. In addition, experiments were conducted with the baseline NMF decoding method, with one parameter variation: the optional use of an extra `filler unit' column in the matrix of slot value activations.
We tested the
system configurations with increasing amounts of training data,
resulting in learning curves. For each speaker, a separate learning
curve was produced, using only that speaker's data. This setup mimics
the ALADIN system in real life: the system is trained progressively
on a particular user's data and adapts itself to the user's language
over time as new phrases or words are introduced over time, by
retraining the system at regular intervals.

A fixed test set was selected for each speaker, and the remaining data
of the speaker were used for training. The original order of the
utterances as they had been recorded was preserved, in order to mimic
the ALADIN training situation, including possible changes in command
structure over time. The test set consisted of the last 20 {\tt movecard}
utterances and the surrounding {\tt dealcard} utterances. We constructed 
the test set around the number of {\tt
  movecard} utterances, as accurate decoding of the {\tt
  movecard} utterances is the most challenging task. 

The remaining training utterances were split into partitions of 25
utterances. For each experiment, the first {\em k} partitions were used for
training, with {\em k} starting at 1 and gradually increasing up to the
maximum number of partitions. The command transcriptions and the
automatically generated action frames were used as training input. For
testing, only command transcriptions were used as input, and the
output consisted of semantic frames induced by the system. The oracle
frames from PATCOR were used as a reference for evaluation (cf. Table \ref{actionFrames})

Each unique experiment, with a unique combination of parameters and
the data of one single speaker, was run ten times, to account for
possible performance differences due to different random system
initializations (for instance at the beginning of the NMF
procedure). The number of HMM training iterations (Baum-Welch) per
experiment was set to twenty. 

As a final evaluation experiment, we observed the best parameters and settings
established during the experiments on the eight users and applied these
to held-out data from an additional user for which more data is
available. The results
of this experiment are presented in Section \ref{evaluation}.

\subsection{Scoring}
\label{scoring}

Scoring was based on a comparison between the semantic frames induced
by the system and the oracle command frames in PATCOR. The used
metrics are the slot precision, recall and
F\textsubscript{\begin{math}\beta=1\end{math}}-score. These metrics 
are commonly used for the evaluation of frame-based systems for spoken
language understanding \cite{Wang2011}. The slot
F\textsubscript{\begin{math}\beta=1\end{math}}-score is 
the harmonic mean of the slot precision and the slot recall. The
following formulas were used for calculation:\\  

slot precision = \# correctly filled slots / \# total filled slots in induced frames\\
\indent slot recall = \# correctly filled slots / \# total filled slots in oracle frames\\
\indent slot F\textsubscript{\begin{math}\beta=1\end{math}}-score = 2 * slot precision * slot recall / (slot precision + slot recall)\\
 
This means that only slots that are filled with a correct value are
rewarded, and both slots that are falsely filled and slots that are
falsely left empty are penalized. When an induced frame is of another
type than the corresponding oracle frame, the filled slots in the
induced frame and in the oracle frame are consequently different,
which automatically results in a relatively large drop in the slot
F-score.  

Various micro-averaged scores were computed, for instance micro-averaged scores across ten different runs (with different random system initializations) of the same experiment, and across experiments with different speakers' data. Computing micro-averaged scores across multiple experiments was carried out by aggregating the slot counts (i.e. number of correctly filled slots and total number of filled slots in induced frames and in oracle frames) of all the included experiments, and calculating the scores based
on these accumulated slot counts, using the aforementioned formulas.

\section{Results \& Discussion}
\label{results}

\begin{table}
\caption{Top-ranked scores with NMF decoding. All scores (Prec. = slot precision, Rec. = slot recall, F = slot F-score) are micro-averaged scores.}
\begin{tabular}{lcccc}\hline\hline
& \textbf{\footnotesize all} & \multicolumn{3}{c}{\textbf{\footnotesize 150 training inst.}}\\\hline
\textbf{\footnotesize Command}& \textbf{\footnotesize F} & \textbf{\footnotesize Prec.} & \textbf{\footnotesize Rec.} & \textbf{F}\\ \hline 
{\footnotesize phoneme uni}	& 20.7& 	13.1& 	28.7 & 18.0\\
{\footnotesize phoneme bi} & 52.9& 	45.4& 70.8 & 55.3\\
{\footnotesize word uni} & 	58.5& 	57.3 & 65.5 & 61.2\\
{\footnotesize word bi} & 	73.6& 	76.6 & 85.6 & 80.8\\\hline
\end{tabular}
\label{topRankedNMF}
\end{table}

We first consider NMF decoding as our baseline, the experimental results of which can be found in Table \ref{topRankedNMF}. The best performing systems all used an extra filler unit column in the matrix V\textsubscript{frames}. As expected, the scores are a lot lower than the scores with HMM decoding for most input types (Table \ref{topRankedHMM}), because NMF is unable to capture the temporal aspects of the commands. The scores with word bigrams, however, are a remarkable exception. Apparently, a sufficient amount of contextual information is included in the word bigrams to enable the NMF procedure to disambiguate between different slot values as accurately as the HMM decoding procedure can. NMF can also be observed to sacrifice precision for recall: this is due to the fact that during NMF decoding, multiple slot values can be activated per command unit, resulting in a relatively large number of filled slots in the induced frames. 

\begin{table}
\caption{Top-ranked scores with HMM decoding for each input type, and the parameter values with which these top-ranked scores were produced (`non' under Fillers means non-shared fillers). All scores (Prec. = slot precision, Rec. = slot recall, F = slot F-score) are micro-averaged scores.}
\begin{tabular}{lcccccccc}\hline\hline
& \textbf{\footnotesize all}        & \multicolumn{3}{c}{\textbf{\footnotesize 150 training inst.}}&\multicolumn{4}{c}{\textbf{\footnotesize Parameter Values (sharing)}}\\\hline
\multicolumn{7}{c}{} & \multicolumn{2}{c}{\textbf{\footnotesize E}}\\\cline{8-9}
\textbf{\footnotesize Command}& \textbf{\footnotesize F} & \textbf{\footnotesize Prec.} & \textbf{\footnotesize Rec.} & \textbf{F} & \textbf{\footnotesize Fillers} & \textbf{\footnotesize T} & \textbf{\footnotesize NMF} & \textbf{\footnotesize HMM}\\ \hline 
{\footnotesize phoneme uni}	& 86.6& 	91.9& 	94.4& 	93.1& 	slot & 	+& 	+& 	+\\
{\footnotesize phoneme bi} & 	86.2& 	91.3& 	95.0& 	93.1& 	non & 	+& 	+& 	-\\
{\footnotesize word uni} & 	88.0& 	93.8& 	90.8& 	92.3& 	slot&	+&+& 	+\\
& 	& 	& 	& 	& 	all&	& 	& 	\\
{\footnotesize word bi} & 	73.0& 	82.1& 	78.5& 	80.3& 	slot&  	-& 	- & -\\
& & 	& 	& 	& 	all&  	& 	+ & \\\hline

\end{tabular}
\label{topRankedHMM}
\end{table}

Table \ref{topRankedHMM} outlines the results of the top-performing HMM
configurations per command input type. 
The system configurations were ranked according to their
overall micro-averaged slot F-scores, which are reported in the first column of 
Table \ref{topRankedHMM}. These scores are based on the induced frames that were 
aggregated across all speakers, training set sizes and experiment
runs (random initializations). The overall slot F-score thus takes into account
the scores at all training set sizes, since the ALADIN application demands for 
steep learning curves, as explained in the Introduction.
The next three columns of Table \ref{topRankedHMM} show the micro-averaged slot 
scores with 150 training utterances -- the largest training set size 
that is shared among all speakers -- for the top-ranked systems. These 
scores were micro-averaged across all speakers and across ten 
experimental runs per speaker. The last four columns show the parameters 
of the top-performing
systems. For each input type, the top row shows the parameter settings
of the system with the highest overall slot F-score. Other parameter values were
added (below the first row) if at least one system with that parameter
value achieved an overall slot F-score that was not significantly
lower than the highest score. Statistical significance of the F-score differences was tested with
approximate randomization testing (as described in \shortcite{Noreen1989}), 
using a critical p-value of 0.05. Only the
scores of the best-performing system are reported for each input type.  

When we look at the scores in Table \ref{topRankedHMM}, we see that
the scores with word bigrams are clearly lower than the scores for
the other input types. This is mainly due to data sparseness: many
unknown word bigrams occur in the test data, resulting in decoding
errors. The overall slot F-scores with phoneme unigrams and phoneme
bigrams are very similar and are not much lower than those with word
unigrams. It seems that the absence of word boundary information in
the input and the smaller command unit size does not have a large impact
on the slot F-scores. The slot F-scores achieved with 150 training
instances are even higher with phoneme unigrams or bigrams than with
word unigrams. However, we do see a difference in the balance between
precision and recall: with phoneme-based command units, recall is
higher than precision, whereas with word-based units, it is the other
way around. The relatively high recall and low precision with
phoneme-based command units can be attributed to the large number of
units per command, which is likely to result in more activated slot
values during decoding. The balance between precision and recall with
word unigrams will be further discussed in subsection
\ref{mostFreqErrors}. 

Looking at the parameter settings of the top-performing HMM-based
systems, in Table \ref{topRankedHMM}, we see some differences between
the optimal settings of the different input types. All top-performing
systems use filler states, but the type of emission probability
sharing they use for the filler states, varies somewhat. All command
input types except phoneme bigrams have top-performing systems that share
the filler state emission probability distributions per slot
({\em slot-shared}). With word-based input, sharing all filler state
emissions produces practically equal results as sharing them per
slot. With phoneme bigrams, on the other hand, the best results are
produced with a system that does not apply any emission sharing to
filler states.  

Regarding the other three parameters -- T-sharing and both types of
E-sharing applied to slot values -- there are also some differences
among the input types. With phoneme or word unigrams as input, the
best results are produced by systems that use all three types of
sharing. With phoneme bigrams, the top-performing system uses
T-sharing and E-sharing in the NMF phase, but no E-sharing in the
HMM. With word bigrams, the top-ranked system uses none of the three
sharing types. The effects of the different parameter settings are
discussed in more detail in the following subsections. In these 
subsections, we will often use the abbreviated slot names (FS, FV, 
etc.), as specified in Fig. \ref{patframes}.

\begin{figure}
\begin{tabular}{c}
{\bf (a) phoneme bigrams}\\
\includegraphics*[width=8cm,trim=0 0 0 40,clip]{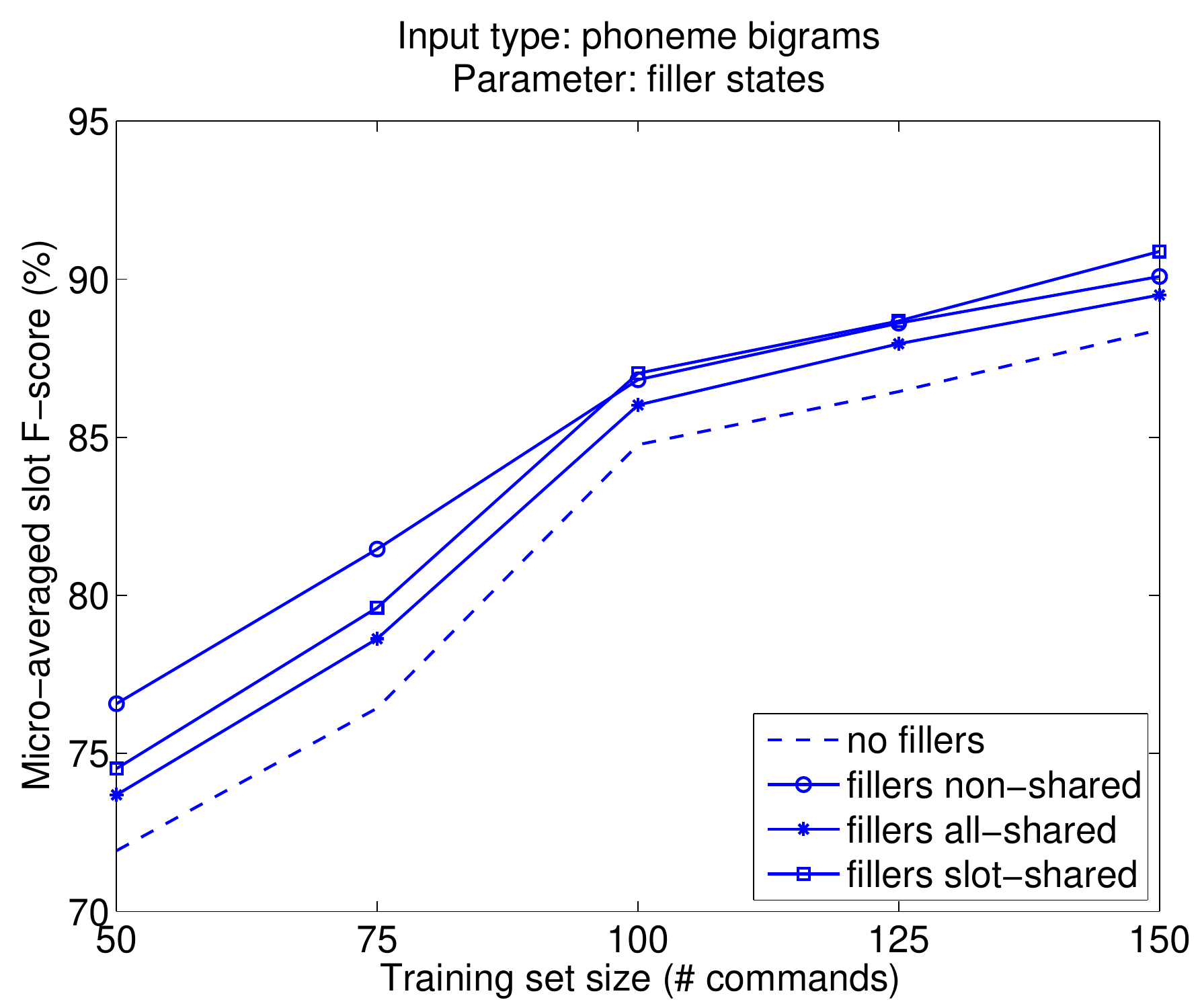}\\
{\bf (b) word unigrams}\\
\includegraphics*[width=8cm,trim=0 0 0 40,clip]{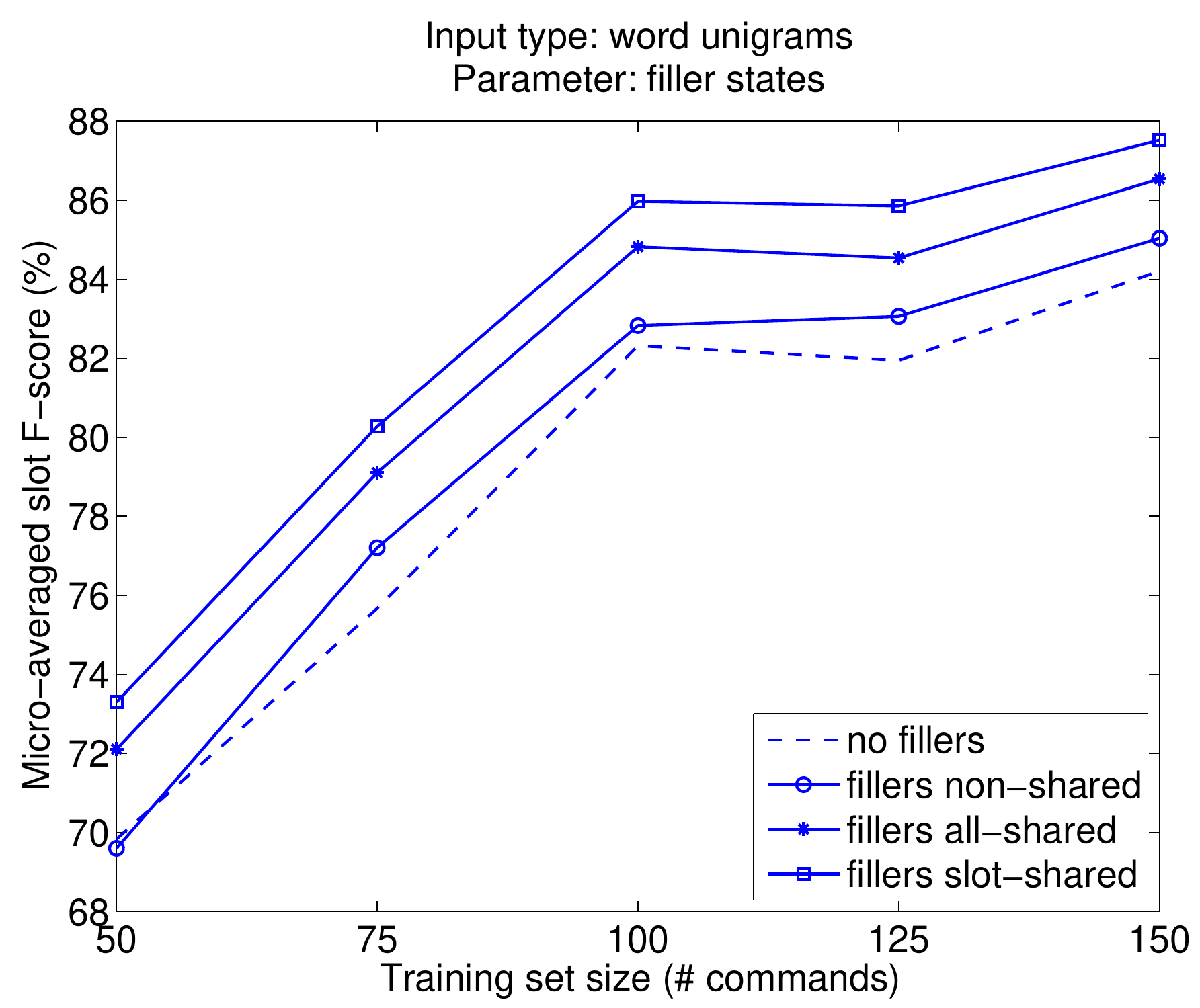}\\
\end{tabular}
\caption{The micro-averaged slot F-scores with different conditions for filler states.}
\label{graphsEffectExtension}
\end{figure}

\subsection{Effects of system extensions}

\begin{figure}
\begin{tabular}{c}
{\bf (a) phoneme bigrams}\\
\includegraphics*[width=8cm,trim=0 0 0 50,clip]{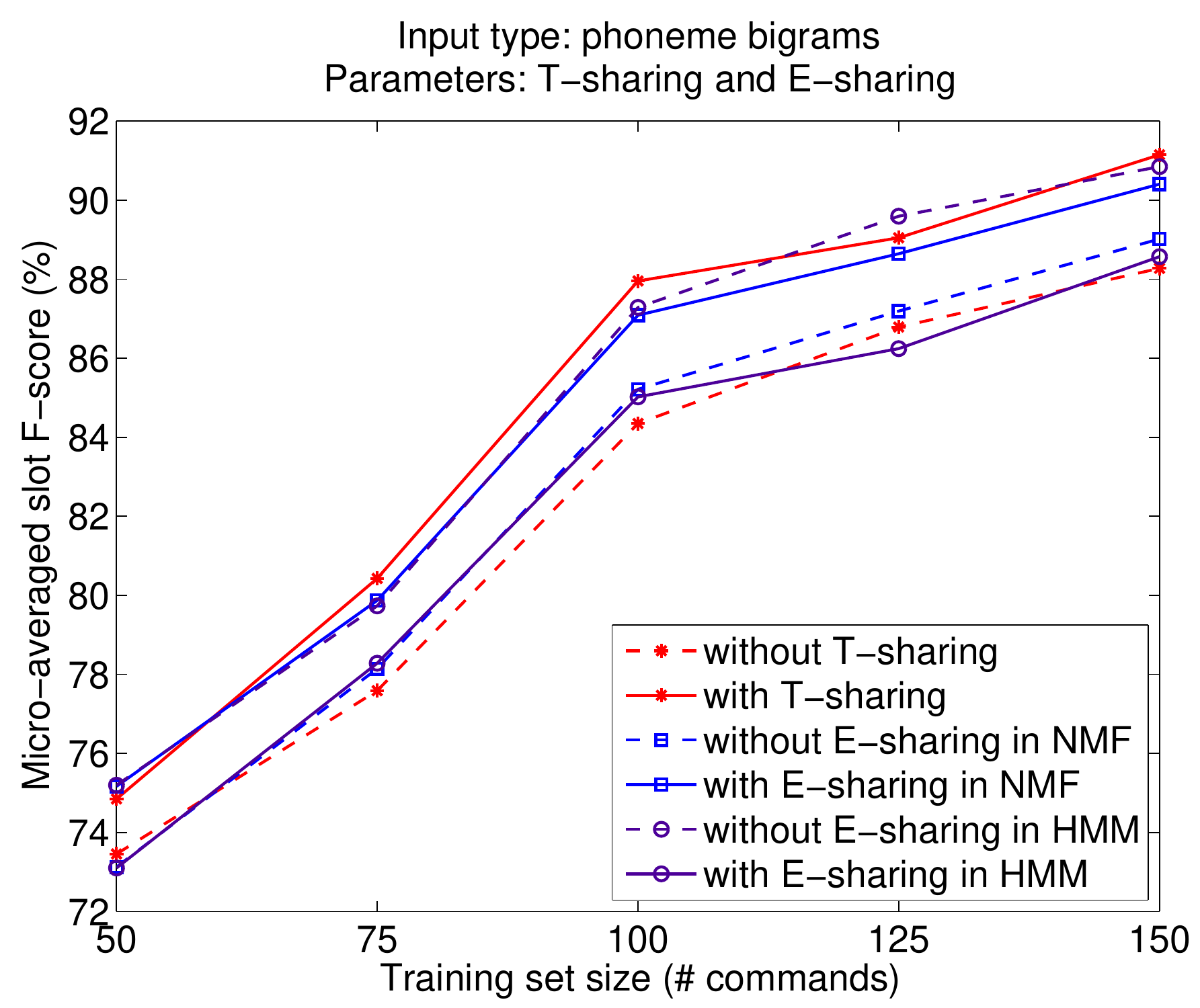}\\
{\bf (b) word unigrams}\\
\includegraphics*[width=8cm,trim=0 0 0 50,clip]{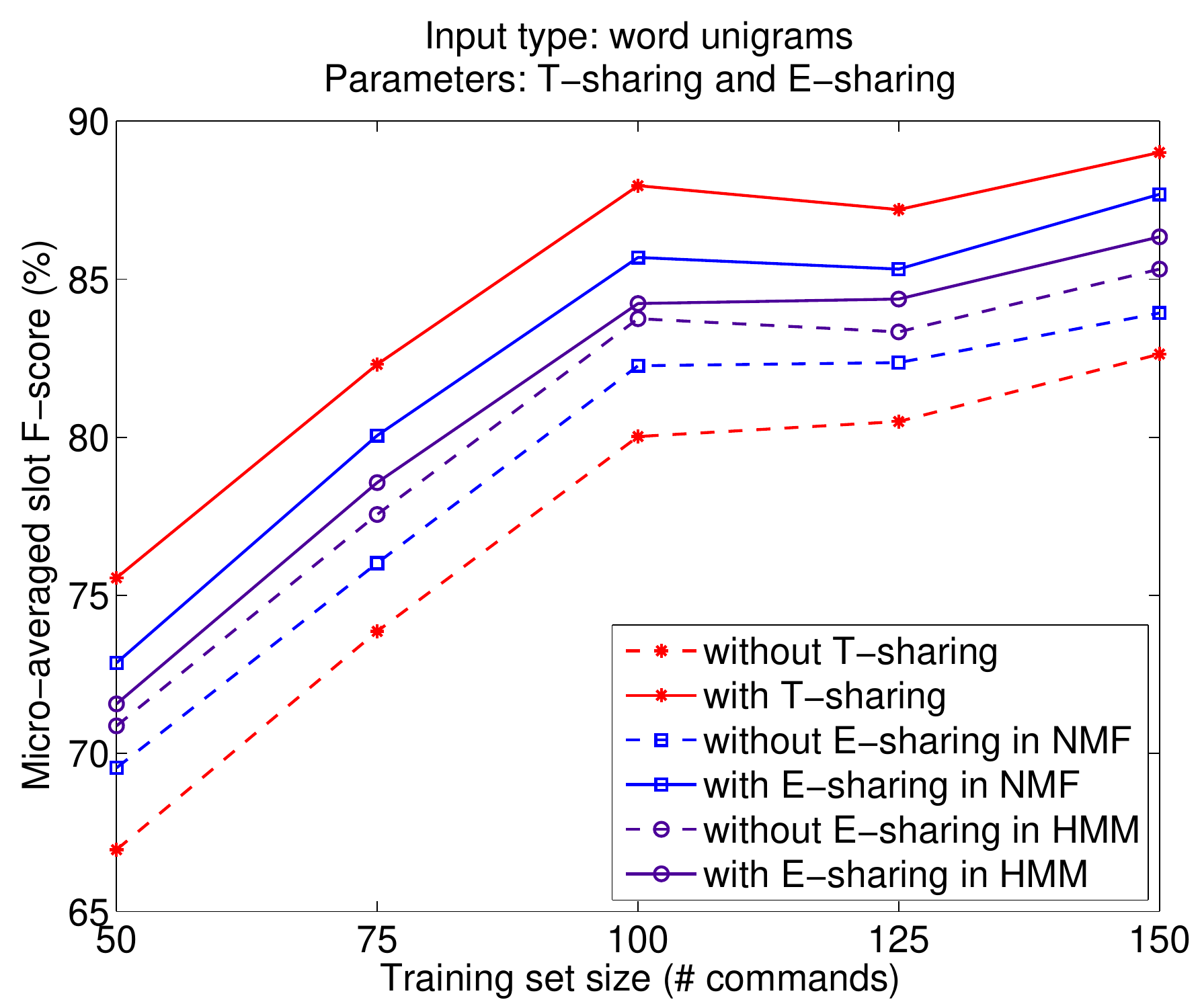}\\
\end{tabular}
\caption{The micro-averaged slot F-scores with different conditions for T-sharing, E-sharing in NMF and in the HMM.}
\label{graphsEffectExtension2}
\end{figure}

In order to compare the parameter effects, the
micro-averaged slot F-scores for the different parameter values were
plotted at different training set sizes.
Figs \ref{graphsEffectExtension} and \ref{graphsEffectExtension2} show the resulting graphs for two input types: phoneme bigrams, which is the input type with the highest recall, and word unigrams, which is the input type with the highest precision (cf. Table \ref{topRankedHMM}).
The micro-averaged F-scores in the graphs
were calculated based on the aggregated set of all semantic frames
that were induced by systems with a specific parameter value. For
instance, the broken lines in Fig. \ref{graphsEffectExtension} show the
F-scores based on all semantic frames that were induced by systems
without filler states (independent of the other parameter values). The
F-scores were thus micro-averaged across all speakers, all system
configurations with a certain parameter value (with different
combinations of other parameter values) and all ten runs per system
configuration. The effects of the different parameter values will be
discussed individually in the following subsections.

The top-performing filler state configurations in Fig. \ref{graphsEffectExtension} correspond to the filler state configurations of the top-performing systems in Table \ref{topRankedHMM}, viz. non-shared filler states for phoneme bigrams, and slot-shared filler states for word unigrams, followed by all-shared filler states. However, in Fig. \ref{graphsEffectExtension} we get a slightly different perspective than in Table \ref{topRankedHMM}. We can see that with phoneme bigrams (Fig. \ref{graphsEffectExtension}(a)), non-shared filler states produce
the best results for smaller training set sizes, while for larger
training set sizes, the slot-shared and non-shared filler states yield
similar top-ranked scores. With word unigrams, 
the slot-shared filler states seem to have a consistent advantage over all-shared filler states when averaging the F-scores across all system configurations  (Fig. \ref{graphsEffectExtension}(b)), while for the specific top-ranked systems in Table \ref{topRankedHMM} (with T-sharing and both types of E-sharing), the overall F-score difference between the system with slot-shared filler states and the one with all-shared filler states was not significant. 

Fig. \ref{graphsEffectExtension2} affirms that T-sharing has a positive
effect for both input types. With word
unigrams, the effect is larger than the effects of the two types of
E-sharing, while with phoneme bigrams, the effect is similar in size to the effect of E-sharing in NMF.
When we look at the resulting slot value sequences, we see
that they are more consistent and accurate regarding
the sequential slot structures they contain, when T-sharing is
used. For instance, the slot sequence ``$<${\tt from\_suit}$>$
$<${\tt from\_value}$>$ $<${\tt target\_suit}$>$ $<${\tt
  target\_value}$>$'' that occurs in a lot of commands is more
consistently present in the decodings.   

In addition, Fig. \ref{graphsEffectExtension2} illustrates that the different types of E-sharing provide mixed results across input types.
For both phoneme bigrams and word
unigrams, E-sharing in the NMF phase has a distinctive positive effect. E-sharing in the HMM training phase, on the other hand, has a negative effect for phoneme bigrams, and its positive effect with word unigrams is relatively small.

Table \ref{effectESh} shows the effects of different E-sharing configurations when optimal T-sharing and filler settings are used (as defined in Table \ref{topRankedHMM}).
Both phoneme and word unigrams benefit from E-sharing in both the NMF and HMM phase. With phoneme and word \textit{bigrams}, however, we see different effects.
With word bigrams, neither of the two E-sharing types has a positive effect, and with phoneme bigrams, E-sharing only has a substantial positive effect when it is applied in the NMF phase exclusively. We will discuss the effects of E-sharing in more detail in the qualitative inspection of the decoding output below.

\begin{figure}[ht]
\center
\includegraphics*[width=10cm]{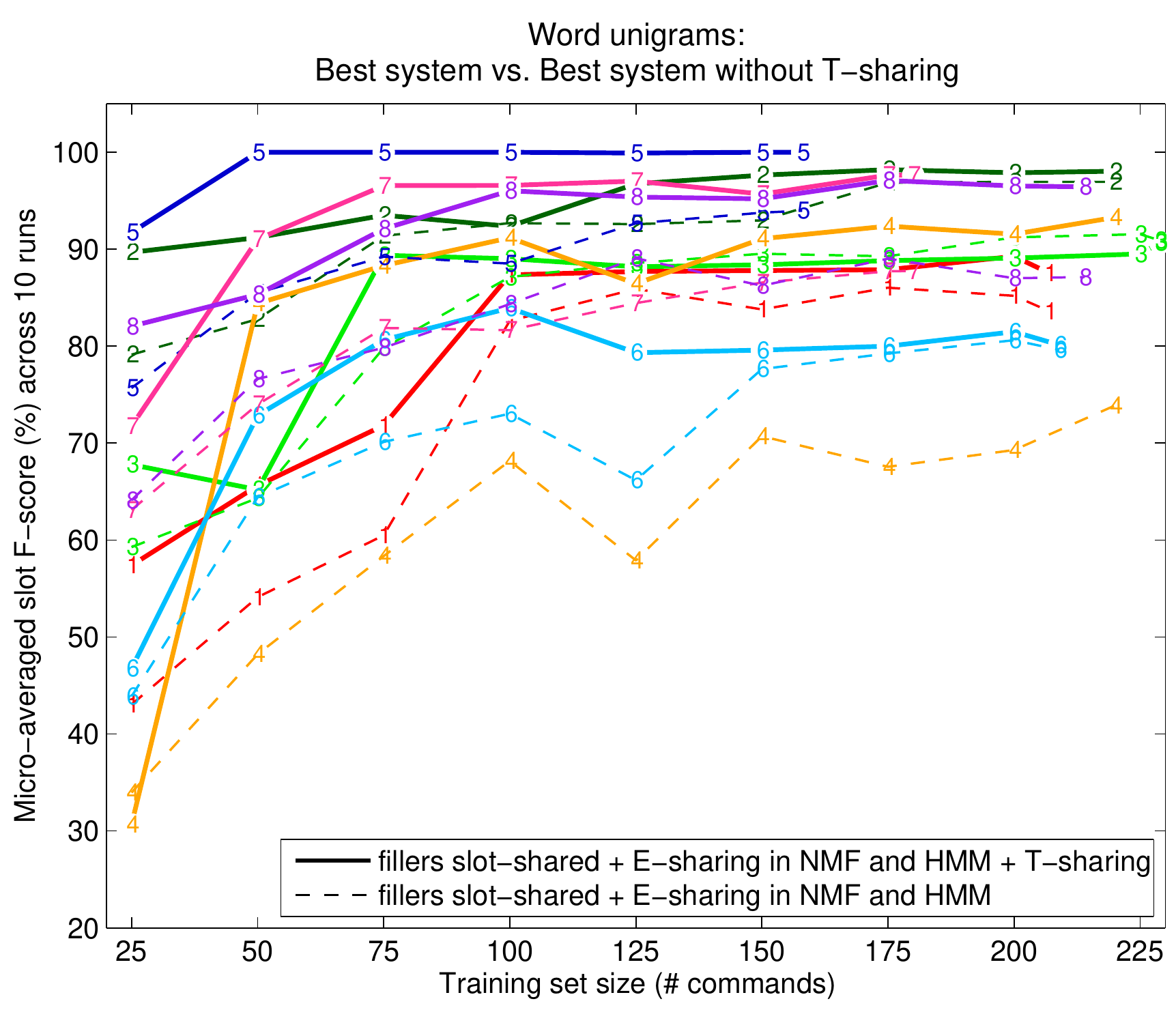}
\caption{Effects of T-sharing on the learning curves of the individual
  speakers with word unigrams as input units. The number markers on
  the curves are the speaker numbers. 
The solid curves show the scores with the best system for word unigrams 
(as specified in Table \ref{topRankedHMM}), which includes T-sharing;
the broken curves show the scores with the same system without T-sharing.}
\label{ppCurvesTSharingEffects}
\end{figure}

\subsubsection*{The effects of T-sharing and E-sharing on individual learning curves}

Fig.~\ref{ppCurvesTSharingEffects} provides some additional insight into
the effect of T-sharing. It displays the learning curves of the
eight speakers with word unigrams as command input for two
system configurations: the top-performing system, in which slot-shared
filler states, both types of E-sharing and T-sharing are used, and the
same system without T-sharing. The effect of T-sharing is substantial,
particularly when dealing with smaller training set sizes. 
For some speakers, the
curves with and without T-sharing converge with larger training set
sizes; for others (speakers 4, 5, 7 and 8), T-sharing keeps showing
considerable positive effects on the F-scores up to the end of the
curves.




\begin{table}
\caption{The effect of E-sharing for the different input types with optimal T-sharing and filler state settings (as defined in Table \ref{topRankedHMM}). Columns 2 through 5 show the overall slot F-scores (micro-averaged across all speakers, training set sizes and experiment runs) with different E-sharing settings.}
\begin{center}
\begin{tabular}{lcccc}\hline\hline
 &\multicolumn{4}{c}{\textbf{E-sharing}}\\ 

\textbf{Input type} & \textbf{none} &\textbf{NMF} & \textbf{HMM} & \textbf{NMF + HMM}\\\hline
phoneme unigrams & 80.05 &	86.05 &	85.50 &	\textbf{86.59}\\
phoneme bigrams & 83.66 & \textbf{86.23}&	84.03&	83.83 \\
word unigrams &	83.23&	86.69&	84.65&	\textbf{88.01}\\
word bigrams & \textbf{73.01} & 	72.75 & 	70.78 & 	71.21 \\\hline
\end{tabular}
\end{center}

\label{effectESh}
\end{table}

\subsubsection*{Qualitative inspection of decoding output}

In order to explain the effects of the different E-sharing types and
the differences between them, we compared the decoding output of
systems with different E-sharing settings and equal T-sharing and
filler state settings. Below, we discuss these results in more detail,
using decoding examples to demonstrate qualitative differences. We will focus
our discussion on word unigrams and phoneme bigrams.

\begin{table}
\caption{The induced slot value sequences for an example sentence at
training size 50 with different types of E-sharing. The first column
  contains the original input, i.e. the phonemic transcription of the
  Dutch utterance (one word per row), the second column contains the
  English translation in orthographic format, and the last four
  columns contain the slot value sequences resulting from the HMM
  decoding process. Decoding errors are marked in bold.}
  
\begin{center}
\begin{tabular}{llrrrr}\hline\hline
& & \multicolumn{4}{c}{\textbf{E-sharing types}}\\
\multicolumn{2}{c}{\textbf{Command}} & \textbf{none} & \textbf{NMF} & \textbf{HMM} & \textbf{NMF + HMM}\\\hline
/d@/ & {\em the} &	filler\_FV=2 &	filler\_FV=2 &	filler\_FV=2 &	filler\_FV=2 \\
/twe/ & {\em two} &	FV=2 &	FV=2 &	FV=2 &	FV=2 \\
/Op/ & {\em on} &	FV=2 &	\textbf{TS=s} &	FV=2 &	filler\_TV=3 \\
/d@/ & {\em the} &	FV=2 &	filler\_TV=3 &	FV=2 &	filler\_TV=3 \\
/dri/ & {\em three} &	\textbf{FV=2} &	TV=3 &	\textbf{FV=2} &	TV=3 \\\hline
\end{tabular}
\end{center}

\label{effectEShWords}
\end{table}

Inspection of the induced slot value sequences revealed that the sequences induced by systems without E-sharing contain many errors pertaining to card values (the slots $<${\tt from\_value}$>$ and $<${\tt target\_value}$>$). An example of such an error is shown in Table \ref{effectEShWords}. 
These decoding errors result from errors in the
earliest stage of the training process: the initial mapping of words
to slot values by NMF. This mapping is impeded by the fact that
subsequent card values typically co-occur in Patience commands (e.g. {\em two
$\rightarrow$ three}, {\em three $\rightarrow$ four}), making it difficult
for the frame induction engine to associate a token with the correct slot value, when only
a limited number of frames and commands have been processed.

A further consequence of the ambiguous card value mappings is that the
sequential command structures are not properly learned either. In other
words: the errors in the initial emission probability distributions
cause errors in the transition probability distributions which are
learned during HMM training. When no E-sharing is used, {\tt from}
to {\tt target} transitions are not favored over {\tt target} to {\tt
  from} transitions. In addition, the probabilities of
self-transitions are strengthened due to the possibility of assigning
the same slot values. This is illustrated in Table \ref{effectEShWords},
where we can observe that without using E-sharing in NMF, the whole sequence is decoded as FV=2, including the word `three'. 

Applying E-sharing in the NMF phase solves the problem of ambiguous
word-to-slot-value mappings by adding extra slot values to the frame
supervision. For each {\tt from} slot value, the corresponding {\tt
  target} slot value is added to the frame supervision, and the other way
around, because {\tt from} slots and their corresponding {\tt target} slots
form shared expression sets. 
The column {\em E-sharing in NMF and HMM} in Table \ref{effectEShWords}
shows that adding E-sharing in
the HMM on top of E-sharing in NMF corrects the remaining
errors in this example. 

The fact that E-sharing in the HMM phase has a smaller positive effect on
the slot F-scores with word unigrams as input type is partly due to the
later stage in which sharing takes place. E-sharing in NMF can make
major differences in the emission and transition probabilities,
because it provides a better starting point for HMM learning, while
E-sharing in the HMM can only regulate the last part of the learning
process. In addition, E-sharing in the HMM phase applies expression sharing
in a more subtle way than E-sharing in the NMF phase. Rather than adding extra
associations between slot values and command units, the association
strengths between {\tt from} and {\tt target} slot values and the command
units that express them, are averaged amongst each other (for instance,
the probabilities of the emissions FV=3 --$>$ {\em three} and TV=3 --$>$
{\em three} are averaged).  

Table~\ref{effectESh} shows that E-sharing has less of a positive effect
with bigram input types. 
This can be explained by the fact that E-sharing can
introduce errors in the mappings between slot values and command units
when bigrams are used as input. 
This is illustrated by the following example, which shows the start of the
command ``harten acht op schoppen negen" (eight of hearts on nine of spades):  

\vspace{0.2cm}
\begin{center}
\begin{tabular}{lll}\hline\hline
\textbf{Command} & \textbf{Original frame}  & \textbf{Additional frame}\\
& \textbf{supervision} & \textbf{supervision (E-sharing)}\\ \hline
/+h/& 	FS=h & 	{\bf TS=h} \\
/hA/& FS=h & TS=h		\\
/Ar/	& FS=h & TS=h	\\
/rt/	& FS=h & TS=h	\\
/t@/& FS=h & TS=h		\\
/@A/& FV=8& TV=8		\\
/Ax/& FV=8& TV=8		\\
/xt/	& FV=8 & TV=8	\\
/tO/ &	FV=8 &	{\bf TV=8} \\
/Op/& Filler & 	Filler	\\
/ps/ &	TS=s &	{\bf FS=s} \\
/sX/& TS=s& FS=s		\\
/XO/& TS=s& FS=s		\\
...	& & 	\\\hline
\end{tabular}
\end{center}
\vspace{0.2cm}
In this example, the slot values in the frame supervision are
aligned with the command units they are likely to be mapped to in the
NMF phase when E-sharing is applied. In this case, the additional
mappings, caused by the frame supervision that is added by applying
E-sharing, are not all correct; see the errors marked in bold. For instance, the unit /ps/ should 
unequivocally express TS=s (due to the presence of the prefiller
{\em op} in the bigram), but is here erroneously marked as FS=s as well.
Such incorrect mappings typically occur with bigrams that cross word boundaries.

This can furthermore explain the E-sharing effects in Fig.~\ref{graphsEffectExtension}(c). With
phoneme bigrams, E-sharing in NMF still has a large positive effect
because of its disambiguation of the slot value mappings, as explained
previously. It also introduces some errors in the slot
value mappings, but only for command units that cross word
boundaries. In addition, these errors can still be corrected in the
HMM training phase, if no E-sharing is applied there. When E-sharing
is applied in the HMM training phase, the same type of mapping errors
are introduced, but in that case, they cannot be corrected
anymore. This also explains why only applying
E-sharing in the NMF phase yields better scores than applying
E-sharing both in the NMF phase and in the HMM training phase, 
as shown in Table \ref{effectESh}.

In summary, applying E-sharing in the NMF phase yields better results
than applying it in the HMM training phase, because applying it at an
early stage allows it to have a relatively large positive effect, and
the possible errors that it introduces -- in case of bigram-based input
-- can still be corrected at a later stage of the learning process. 

\subsection{Most frequent errors with optimal settings}
\label{mostFreqErrors}

We analyzed the remaining errors that occurred with the optimal
settings and concentrate
on the most frequent errors with word unigrams as input type at 150
training utterances. As can be seen in Table \ref{topRankedHMM}, the
recall was lower than the precision (90.78\% vs. 93.84\%) when word
unigrams were used. The most frequent error, which had a negative
effect on the recall, was that utterances such as ``{\em aas naar
  boven}", in which the ace was moved to one of the
foundation stacks, were often tagged as sequences of one single
repeated slot value: either FV=1 or some TF value. This
error occurred with almost all speakers and is due to the fact that
the ace is always moved to a foundation stack and not in the playing field below.

Other errors that occurred frequently are incorrectly
tagged interjections such as `uh', `ja', etc. Sometimes these errors
also percolated to other parts of the utterances. In some cases,
the interjection was an unknown word, and the error could have been
prevented by ignoring the word instead of mapping it onto a similar
known word, as we did now. Apart from short interjections, other
disfluencies such as restarts were amongst the main  causes for error.

\section{Evaluation experiments with optimal settings}
\label{evaluation}

We carried out evaluation experiments with the optimal settings that
were established in the previous experiments. The optimal settings are
shown in Table \ref{topRankedHMM}. For these experiments, we used
speaker 9's data subset, which was not used in the previous
experiments and is much larger than the data subsets of speakers 1
through 8. It consists of 1,142 commands and corresponding semantic
frames. The last 200 {\tt movecard} commands and the surrounding 211
{\tt dealcard} commands were used as a test set, and the remaining
commands -- 440 {\tt movecard} commands and 291 {\tt dealcard}
commands -- were used for training. As in the previous experiments,
the training set was divided into partitions of 25 utterances, and
increasing numbers of partitions were used for training in order to
produce learning curves. For word bigrams, we carried out evaluation experiments with two system configurations: the overall optimal configuration -- using NMF decoding instead of using an HMM -- as well as the optimal configuration with the use of an HMM (see Table \ref{topRankedHMM}). All experiments were
carried out ten times, and micro-averaged precision, recall and
F-scores were calculated in the same way as in the previous
experiments. 

\begin{figure}
\begin{tabular}{c}
{\bf (a)} \\
\includegraphics[width=8cm]{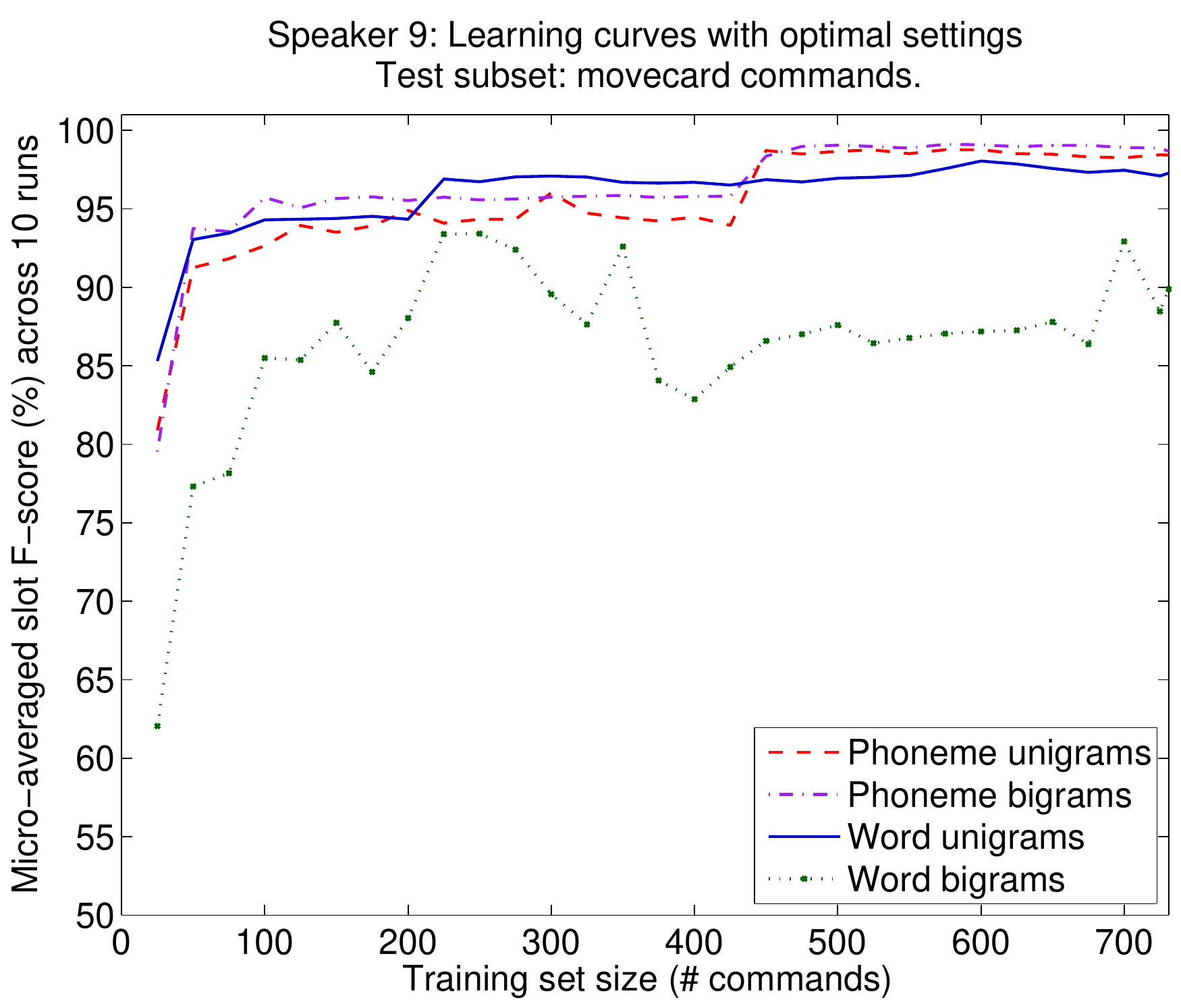}\\
{\bf (b)} \\
\includegraphics[width=8cm]{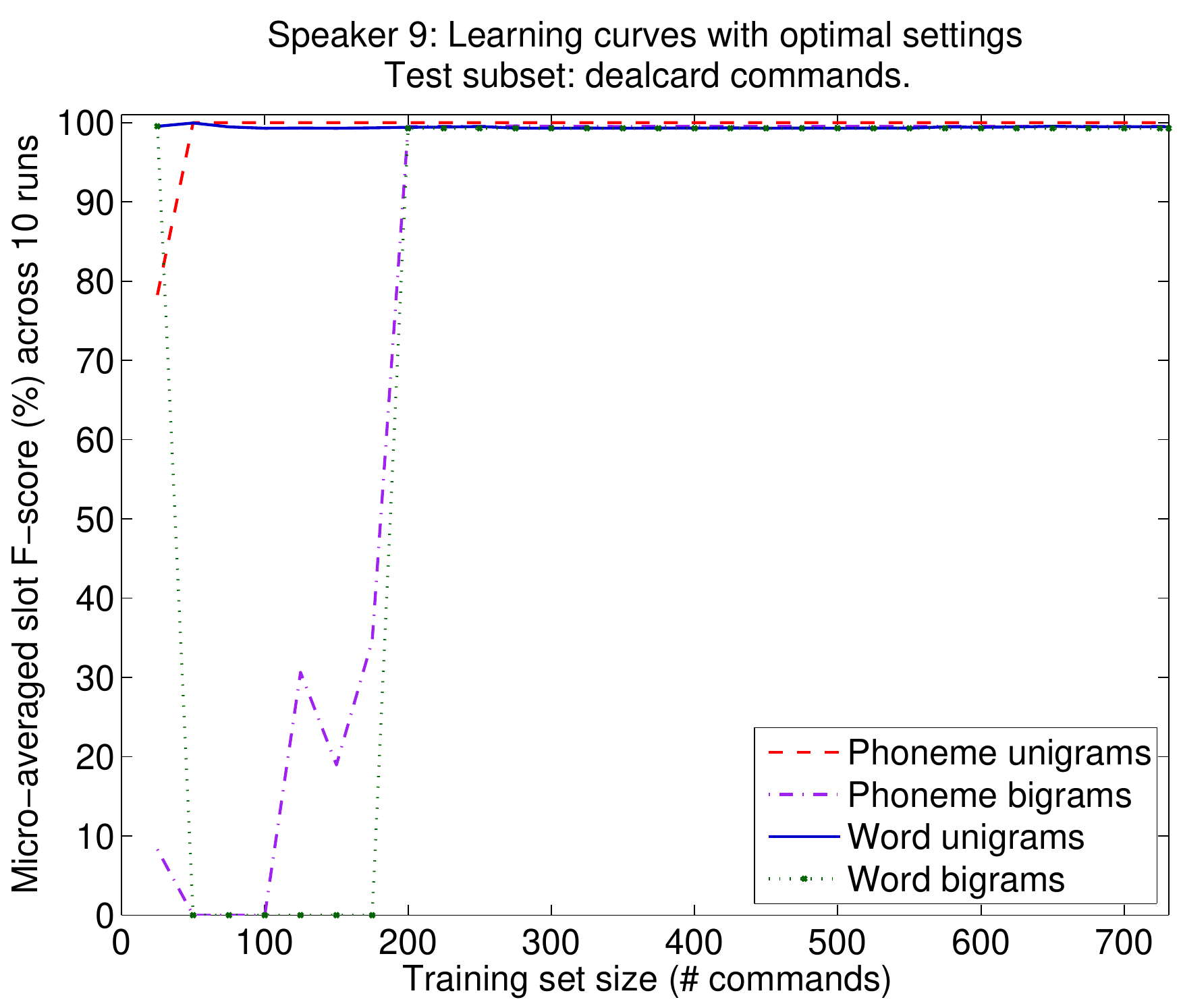}\\
\end{tabular}
\caption{Learning curves resulting from evaluation experiments with
  speaker 9's data, using the optimal parameter settings for each
  input type. The scores for the {\tt movecard} commands are shown in graph (a); the scores for the {\tt dealcard} commands are shown in graph (b).}
\label{EvalLearningCurves}
\end{figure}

Figure \ref{EvalLearningCurves} shows the resulting learning
curves for {\tt movecard} commands (Fig. \ref{EvalLearningCurves}(a)) and {\tt
  dealcard} commands (Fig. \ref{EvalLearningCurves}(b)). With phoneme
unigrams or bigrams and word unigrams, the F-scores for {\tt movecard}
commands are already between 90\% and 95\% with only 50 training
commands (which is an average game of Patience). These
curves start to level off at 100 training commands, between 94\% and
95\%, but the errors that are made are mainly due to the fact that the
word {\em heer} (king), which appears quite regularly in the test set, only
starts to appear in the training set after the 200th command (until
then, the speaker uses the synonym {\em koning} instead). With word
unigrams, the first appearances of {\em heer} in the training set directly
result in a leap up to approximately 97\%. With phoneme unigrams and
bigrams, the leap appears later, at 450 training commands, once the
word {\em heer} has appeared in both the {\tt from} and the {\tt target}
position. That leap results in F-scores of around 99\%, which is a bit
higher than the maximum score that is reached with word unigrams
(around 98\%, above 500 training commands). 
With word bigrams, the
scores are lower than with the other input types and simple NMF decoding mostly outperforms HMM decoding. This
corresponds to the scores we saw in the previous experiments. Once the
word {\em heer} starts to appear in the training data (at training size
225), the scores go up to about 93\%. 

In Figure \ref{EvalLearningCurves}(b), the learning
curves for the {\tt dealcard} commands are shown. With phoneme and
word unigrams, the F-scores for {\tt dealcard} commands are
approximately 100\% even with small training set sizes. With bigrams,
however, the F-scores are lower for training set sizes below 200
commands. This is caused by the fact that in those smaller training
sets, {\tt dealcard} utterances are always ``{\em nieuwe kaarten
omdraaien}'', while in many {\tt dealcard} commands after that, the
word ``{\tt nieuwe}'' is omitted. In the {\tt dealcard} commands in the test
set, the word ``{\tt nieuwe}'' is omitted as well.

\section{Conclusion \& Future Work}
\label{conclusion}

Our contribution to the state-of-the-art in the field of semantic frame
induction, is threefold. Firstly, we presented a new application
context for the task of weakly supervised semantic frame induction,
viz. a speaker-dependent vocal interface geared towards physically impaired
users that automatically learns a user's specific pronunciation,
vocabulary and command structure from a small set of commands and
associated controls. The weak supervision consists of
automatically generated semantic frames, of which the slots are not
aligned with segments in the commands, and which usually contain redundant
information that is not expressed in the commands. Secondly, we
described a framework based on NMF and HMM learning to complete this
task, and some system extensions to improve its performance: HMM structure
extensions and expression sharing. Unlike most SLU systems, our system directly induces frame slots \textit{and} their values, while the HMM structure extensions keep the parameter space manageable. Thirdly, we presented a detailed
analysis of the effects of the system extensions, based on textual
command input (transcriptions). The used corpus is PATCOR,
which contains Dutch-spoken commands in the context
of a voice-controlled card game. Apart from using command input based
on word unigrams, as is usually done in SLU research, we also
experimented with word bigrams, phoneme unigrams and phoneme bigrams
as observed command units. 

In general, the results show positive effects of all the described
system extensions. Sharing transition probabilities, resulting in a \textit{hierarchical} HMM, has a
considerable positive effect on the learning speed with all input
types except word bigrams. The addition of filler states to deal with
words that do not express slot values also shows positive effects, and
the best results were produced with filler states that shared their
emission probabilities slotwise. Expression sharing in the NMF phase
has a positive effect for all input types except word bigrams,
whereas expression sharing in the HMM phase only has positive effects
with (word or phoneme) unigrams as input. 
The more positive effect
of expression sharing in the NMF phase is mainly due to its early
application in the learning process, which makes the improvement
(viz. with unigram input) relatively large and enables the system to
correct possible negative effects (viz. with bigram input) in a later
learning stage. 

Finally, evaluation experiments with the
top-performing system configurations on held-out data show very encouraging 
learning results. With word unigrams, phoneme unigrams and phoneme bigrams
as input types, scores
between 90\% and 95\% can already be achieved with only 50 training
utterances, and the main error cause was a late shift in the
speaker's vocabulary. With larger training sets, in which this
inconsistency was resolved, F-scores up to 98\% (with word unigrams)
and 99\% (with phoneme unigrams and bigrams) were achieved, further underlining
the system's ability to adapt to changes in language use over time.

In future research, we plan to evaluate 
our extended ALADIN semantic frame induction system on other datasets. The
scene description task in \shortcite{Roy2002} involves learning
words and syntax on the basis of redundant sets of visual features. Similarly,
the Robocup Sportscasting dataset \cite{Chen2008} contains utterances of
humans commenting on simulated Robocup soccer games, coupled with (redundant)
semantic descriptions of the scenes. 
The latter dataset has received a lot
of attention with previous research efforts focusing on aligning utterances
with frames \cite{Liang2009} and learning semantic parsers \cite{Chen2008,Chen2010,Kim2010,Borschinger2011}.
The ALADIN approach may provide
an interesting addition to the state-of-the-art for this dataset, as it offers a relatively straightforward framework for semantic frame slot filling on the basis
of utterances.

In the context of the ALADIN project, we will perform additional experiments 
in which we use the output of the frame induction engine 
presented in this work, as training data for a discriminative concept tagger \cite{nlp4ita}. 
Experiments show that this post-processing step further improves 
F-scores, as this type of concept tagging is able to take more context into account during classification. 
Finally, we will 
also evaluate performance of the frame induction technique and its extensions
using acoustic units as input type, as well as experiments on a home automation 
dataset containing both non-pathological and pathological speech.

\section*{Acknowledgments}

This research described in this paper was funded by IWT-SBO grant 100049 (ALADIN). The PATCOR dataset is available at \url{https://github.com/clips/patcor}.

\bibliographystyle{apalike}
\bibliography{janneke}

\begin{thebibliography}{}

\bibitem[Baum, 1972]{Baum1972}
Baum, L.~E. (1972).
\newblock {An equality and associated maximization technique in statistical
  estimation for probabilistic functions of markov processes}.
\newblock {\em Inequalities}, 3:1--8.

\bibitem[Bonneau-Maynard et~al., 2009]{Bonneau2009}
Bonneau-Maynard, H., Quignard, M., and Denis, A. (2009).
\newblock Media: a semantically annotated corpus of task oriented dialogs in
  french.
\newblock {\em Language Resources and Evaluation}, 43(4):329--354.

\bibitem[B{\"o}rschinger et~al., 2011]{Borschinger2011}
B{\"o}rschinger, B., Jones, B.~K., and Johnson, M. (2011).
\newblock Reducing grounded learning tasks to grammatical inference.
\newblock In {\em Proceedings of the Conference on Empirical Methods in Natural
  Language Processing}, pages 1416--1425. Association for Computational
  Linguistics.

\bibitem[Chang, 1993]{Chang1993}
Chang, H.-P. (1993).
\newblock Speech input for dysarthric users.
\newblock {\em The Journal of the Acoustical Society of America},
  94(3):1782--1782.

\bibitem[Chen et~al., 2010]{Chen2010}
Chen, D.~L., Kim, J., and Mooney, R.~J. (2010).
\newblock Training a multilingual sportscaster: Using perceptual context to
  learn language.
\newblock {\em Journal of Artificial Intelligence Research}, 37(1):397--436.

\bibitem[Chen and Mooney, 2008]{Chen2008}
Chen, D.~L. and Mooney, R.~J. (2008).
\newblock Learning to sportscast: a test of grounded language acquisition.
\newblock In {\em Proceedings of the 25th international conference on Machine
  learning}, pages 128--135. ACM.

\bibitem[Chen and Mooney, 2011]{Chen:2011:LIN:2900423.2900560}
Chen, D.~L. and Mooney, R.~J. (2011).
\newblock Learning to interpret natural language navigation instructions from
  observations.
\newblock In {\em Proceedings of the Twenty-Fifth AAAI Conference on Artificial
  Intelligence}, AAAI'11, pages 859--865. AAAI Press.

\bibitem[Dahl et~al., 1994]{Dahl1994}
Dahl, D.~A., Bates, M., Brown, M., Fisher, W., Hunicke-Smith, K., Pallett, D.,
  Pao, C., Rudnicky, A., and Shriberg, E. (1994).
\newblock Expanding the scope of the atis task: The atis-3 corpus.
\newblock In {\em Proceedings of the Workshop on Human Language Technology},
  HLT '94, pages 43--48, Stroudsburg, PA, USA. Association for Computational
  Linguistics.

\bibitem[Della~Pietra et~al., 1997]{Pietra1997}
Della~Pietra, S., Epstein, M., Roukos, S., and Ward, T. (1997).
\newblock Fertility models for statistical natural language understanding.
\newblock In {\em Proceedings of the 35th Annual Meeting of the Association for
  Computational Linguistics}, pages 168--173.

\bibitem[Dempster et~al., 1977]{Dempster1977}
Dempster, A.~P., Laird, N.~M., Rubin, D.~B., et~al. (1977).
\newblock Maximum likelihood from incomplete data via the em algorithm.
\newblock {\em Journal of the Royal statistical Society}, 39(1):1--38.

\bibitem[Dinarelli et~al., 2009]{Dinarelli2009}
Dinarelli, M., Quarteroni, S., Tonelli, S., Moschitti, A., and Riccardi, G.
  (2009).
\newblock Annotating spoken dialogs: from speech segments to dialog acts and
  frame semantics.
\newblock In {\em Proceedings of the 2nd Workshop on Semantic Representation of
  Spoken Language}, pages 34--41. Association for Computational Linguistics.

\bibitem[Elffers et~al., 2005]{Elffers2005}
Elffers, B., Van~Bael, C., and Strik, H. (2005).
\newblock Adapt: Algorithm for dynamic alignment of phonetic transcriptions.
\newblock {\em manual available online from http://lands. let. ru.
  nl/literature/elffers. 2005.1. pdf}.

\bibitem[Epstein et~al., 1996]{Epstein1996}
Epstein, M., Ward, T., Della~Pietra, S., Papineni, K., and Roukos, S. (1996).
\newblock Statistical natural language understanding using hidden clumpings.
\newblock In {\em IEEE International Conference on Acoustics, Speech, and
  Signal Processing 1996 (ICASSP-96)}, volume~1, pages 176--179. IEEE.

\bibitem[Goldwasser and Roth, 2014]{Goldwasser:2014:LNI:2583611.2583673}
Goldwasser, D. and Roth, D. (2014).
\newblock Learning from natural instructions.
\newblock {\em Mach. Learn.}, 94(2):205--232.

\bibitem[Hahn et~al., 2011]{Hahn2011}
Hahn, S., Dinarelli, M., Raymond, C., Lefevre, F., Lehnen, P., de~Mori, R.,
  Moschitti, A., Ney, H., and Riccardi, G. (2011).
\newblock Comparing stochastic approaches to spoken language understanding in
  multiple languages.
\newblock {\em IEEE Transactions on Audio, Speech \& Language Processing},
  19(6):1569--1583.

\bibitem[Hawley et~al., 2007]{Hawley2007}
Hawley, M.~S., Enderby, P., Green, P., Cunningham, S., Brownsell, S.,
  Carmichael, J., Parker, M., Hatzis, A., O'Neill, P., and Palmer, R. (2007).
\newblock A speech-controlled environmental control system for people with
  severe dysarthria.
\newblock {\em Medical Engineering \& Physics}, 5(29):586 -- 593.

\bibitem[Hemphill et~al., 1990]{Hemphill1990}
Hemphill, C.~T., Godfrey, J.~J., and Doddington, G.~R. (1990).
\newblock The atis spoken language systems pilot corpus.
\newblock In {\em Proceedings of the Workshop on Speech and Natural Language},
  HLT '90, pages 96--101, Stroudsburg, PA, USA. Association for Computational
  Linguistics.

\bibitem[Kim and Mooney, 2010]{Kim2010}
Kim, J. and Mooney, R.~J. (2010).
\newblock Generative alignment and semantic parsing for learning from ambiguous
  supervision.
\newblock In {\em Proceedings of the 23rd International Conference on
  Computational Linguistics: Posters}, pages 543--551. Association for
  Computational Linguistics.

\bibitem[Lee and Seung, 1999]{lee1999}
Lee, D.~D. and Seung, H.~S. (1999).
\newblock Learning the parts of objects by non-negative matrix factorization.
\newblock {\em Nature}, 401(6755):788--791.

\bibitem[Liang et~al., 2009]{Liang2009}
Liang, P., Jordan, M.~I., and Klein, D. (2009).
\newblock Learning semantic correspondences with less supervision.
\newblock In {\em Proceedings of the Joint Conference of the 47th Annual
  Meeting of the ACL and the 4th International Joint Conference on Natural
  Language Processing of the AFNLP: Volume 1 - Volume 1}, ACL '09, pages
  91--99, Stroudsburg, PA, USA. Association for Computational Linguistics.

\bibitem[Macherey et~al., 2001]{Macherey2001}
Macherey, K., Och, F.~J., and Ney, H. (2001).
\newblock Natural language understanding using statistical machine translation.
\newblock In {\em INTERSPEECH}, pages 2205--2208.

\bibitem[Mertens and Vercammen, 1998]{Mertens1998}
Mertens, P. and Vercammen, F. (1998).
\newblock Fonilex manual.
\newblock Technical report, K.U.Leuven - CCL.

\bibitem[Mykowiecka et~al., 2009]{Mykowiecka2009}
Mykowiecka, A., Marasek, K., Marciniak, M., Rabiega-Wisniewska, J., and
  Gubrynowicz, R. (2009).
\newblock Annotated corpus of polish spoken dialogues.
\newblock In Vetulani, Z. and Uszkoreit, H., editors, {\em Human Language
  Technology. Challenges of the Information Society}, volume 5603 of {\em
  Lecture Notes in Computer Science}, pages 50--62. Springer Berlin Heidelberg.

\bibitem[Noreen, 1989]{Noreen1989}
Noreen, E.~W. (1989).
\newblock {\em Computer intensive methods for testing hypotheses: an
  introduction}.
\newblock Wiley, New York.

\bibitem[Ons et~al., 2014]{Ons2014997}
Ons, B., Gemmeke, J.~F., and {Van hamme}, H. (2014).
\newblock Fast vocabulary acquisition in an nmf-based self-learning vocal user
  interface.
\newblock {\em Computer Speech \& Language}, 28(4):997 -- 1017.

\bibitem[Ons et~al., 2013]{ons2013self}
Ons, B., Tessema, N., van~de Loo, J., and Gemmeke, J.~F. (2013).
\newblock A self learning vocal interface for speech-impaired users.
\newblock {\em Proceedings SLPAT 2013}, pages 1--9.

\bibitem[Oostdijk, 2000]{Oostdijk2000}
Oostdijk, N. (2000).
\newblock The spoken dutch corpus. overview and first evaluation.
\newblock In {\em Proceedings of Second International Conference on Language
  Resources and Evaluation (LREC)}, pages 887?--894.

\bibitem[Pieraccini et~al., 1991]{Pieraccini1991}
Pieraccini, R., Levin, E., and Lee, C.-H. (1991).
\newblock Stochastic representation of conceptual structure in the atis task.
\newblock In {\em HLT}, pages 121--124. Morgan Kaufmann.

\bibitem[Roy, 2002]{Roy2002}
Roy, D.~K. (2002).
\newblock Learning visually grounded words and syntax for a scene description
  task.
\newblock {\em Computer Speech \& Language}, 16(3):353--385.

\bibitem[{van de Loo} et~al., 2012]{nlp4ita}
{van de Loo}, J., {De Pauw}, G., Gemmeke, J., Karsmakers, P., {Van Den Broeck},
  B., Daelemans, W., and {Van hamme}, H. (2012).
\newblock Towards shallow grammar induction for an adaptive assistive vocal
  interface: a concept tagging approach.
\newblock In {\em Proceedings NLP4ITA}, pages 27--34.

\bibitem[Van~hamme, 2008]{hugo_hac}
Van~hamme, H. (2008).
\newblock Hac-models: a novel approach to continuous speech recognition.
\newblock In {\em Proceedings INTERSPEECH}, pages 2554--2557.

\bibitem[Viterbi, 1967]{Viterbi1967}
Viterbi, A.~J. (1967).
\newblock Error bounds for convolutional codes and an asymptotically optimum
  decoding algorithm.
\newblock {\em Information Theory, IEEE Transactions on}, 13(2):260--269.

\bibitem[Wang and Acero, 2006]{Wang2006}
Wang, Y. and Acero, A. (2006).
\newblock Rapid development of spoken language understanding grammars.
\newblock {\em Speech Communication}, 48(3-4):390--416.

\bibitem[Wang et~al., 2011]{Wang2011}
Wang, Y., Deng, L., and Acero, A. (2011).
\newblock Semantic frame-based spoken language understanding.
\newblock In Tur, G. and Mori, R.~D., editors, {\em Spoken Language
  Understanding: Systems for Extracting Semantic Information from Speech},
  chapter~3, pages 41--91. Wiley, West-Sussex, UK.

\bibitem[Zettlemoyer and Collins, 2005]{Zettlemoyer:2005:LMS:3020336.3020416}
Zettlemoyer, L.~S. and Collins, M. (2005).
\newblock Learning to map sentences to logical form: Structured classification
  with probabilistic categorial grammars.
\newblock In {\em Proceedings of the Twenty-First Conference on Uncertainty in
  Artificial Intelligence}, UAI'05, pages 658--666, Arlington, Virginia, United
  States. AUAI Press.

\end{thebibliography}

\label{lastpage}

\end{document}